\documentclass[journal]{IEEEtran}
\usepackage{amsmath,amsfonts}
\usepackage{algorithmic}
\usepackage{algorithm}
\usepackage{array}
\usepackage[caption=false,font=normalsize,labelfont=sf,textfont=sf]{subfig}
\usepackage{textcomp}
\usepackage{stfloats}
\usepackage{url}
\usepackage{verbatim}
\usepackage{graphicx}
\usepackage{cite}
\usepackage{xcolor}

\usepackage{multirow}
\usepackage{algorithm}
\usepackage{algorithmic}
\usepackage[T1]{fontenc}
\usepackage{multicol}
\usepackage{stfloats}
\usepackage[caption = false]{subfig}
\usepackage{xcolor}

\usepackage{mathtools}

\usepackage{makecell}
\usepackage{scalerel}
\usepackage{array}
\usepackage{dsfont}
\usepackage{url}
\usepackage{pifont}

\newcolumntype{C}[1]{>{\centering\let\newline\\\arraybackslash\hspace{0pt}}p{#1}}
\newcolumntype{R}[1]{>{\raggedleft\let\newline\\\arraybackslash\hspace{0pt}}p{#1}}
\newcolumntype{L}[1]{>{\raggedright\let\newline\\\arraybackslash\hspace{0pt}}p{#1}}
\newcolumntype{M}[1]{>{\centering\let\newline\\\arraybackslash\hspace{0pt}}m{#1}}
\newcommand*\rot{\rotatebox{90}}
\newcommand\Tstrut{\rule{-3pt}{2.6ex}}       
\newcommand\Bstrut{\rule[-0.9ex]{-3pt}{0pt}} 
\newcommand{\TBstrut}{\rule{-3pt}{2.6ex} \rule[-0.9ex]{-2pt}{0pt}}  

\newcommand\mydots{\hbox to 1em{.\hss.\hss.}}

\definecolor{myblue}{RGB}{0, 250, 0}
\definecolor{mypink}{RGB}{237, 2, 140}

\begin{document}

\title{The Foreseeable Future: Self-Supervised Learning to Predict Dynamic Scenes for Indoor Navigation}


\author{
Hugues Thomas,~\IEEEmembership{Member,~IEEE,}
Jian Zhang,~\IEEEmembership{Member,~IEEE,}
Timothy D. Barfoot,~\IEEEmembership{Fellow,~IEEE}

\thanks{Hugues Thomas and Timothy D. Barfoot are with the Institute for Aerospace Studies (UTIAS), University of Toronto, Canada. Jian Zhang is with Apple, Cupertino, USA.}

\thanks{Manuscript submitted July 28, 2022; regular submission.}
}


\maketitle


\begin{abstract}

We present a method for generating, predicting, and using Spatiotemporal Occupancy Grid Maps (SOGM), which embed future semantic information of real dynamic scenes. We present an auto-labeling process that creates SOGMs from noisy real navigation data. We use a 3D-2D feedforward architecture, trained to predict the future time steps of SOGMs, given 3D lidar frames as input. Our pipeline is entirely self-supervised, thus enabling lifelong learning for real robots. The network is composed of a 3D back-end that extracts rich features and enables the semantic segmentation of the lidar frames, and a 2D front-end that predicts the future information embedded in the SOGM representation, potentially capturing the complexities and uncertainties of real-world multi-agent, multi-future interactions. We also design a navigation system that uses these predicted SOGMs within planning, after they have been transformed into Spatiotemporal Risk Maps (SRMs). We verify our navigation system's abilities in simulation, validate it on a real robot, study SOGM predictions on real data in various circumstances, and provide a novel indoor 3D lidar dataset, collected during our experiments, which includes our automated annotations.

\end{abstract}






%
%
%
%
%
%
%
%
%
%
%
%

\section{Introduction}

\IEEEPARstart{P}{redicting} the future has always fascinated humanity. From the Oracle of Delphi to Paul the Octopus, this curiosity for the unknown has never faded. But we tend to forget that we already predict the future constantly in our daily lives, only it is for a short horizon. Walking in the street, catching a falling object, or driving a car, all these actions require a certain level of anticipation. With practice, humans can become quite good at predicting what might happen for the next few seconds in many situations; what about robots?

\begin{figure}[!t]
\centering
\includegraphics[width=\columnwidth, keepaspectratio=true]{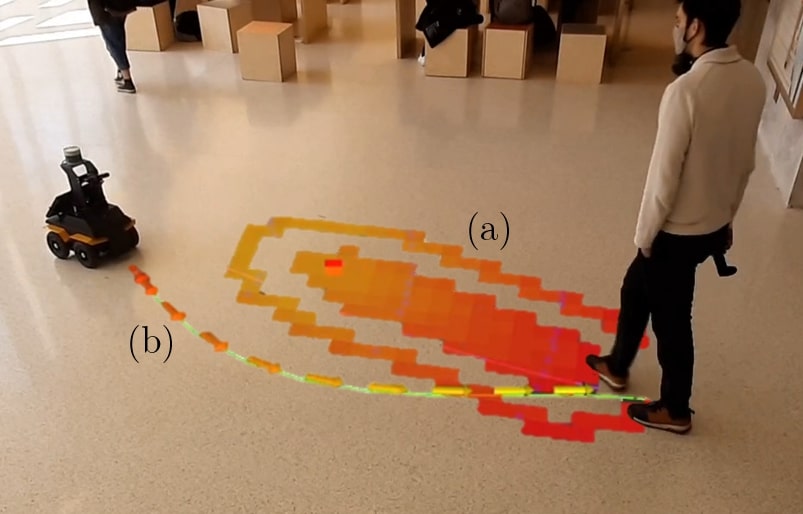}
\caption{Our robot navigating in a real dynamic scene. Future occupied locations are predicted as  Spatiotemporal Occupancy Grid Maps (a) and the robot plans a trajectory to avoid them (b). Time is represented as a color, from red (now) to yellow (future). The ring and center areas represent low and high occupancy probabilities respectively.}
\label{fig_intro}
\end{figure}

\begin{figure*}[t]
\centering
\includegraphics[width=\textwidth, keepaspectratio=true]{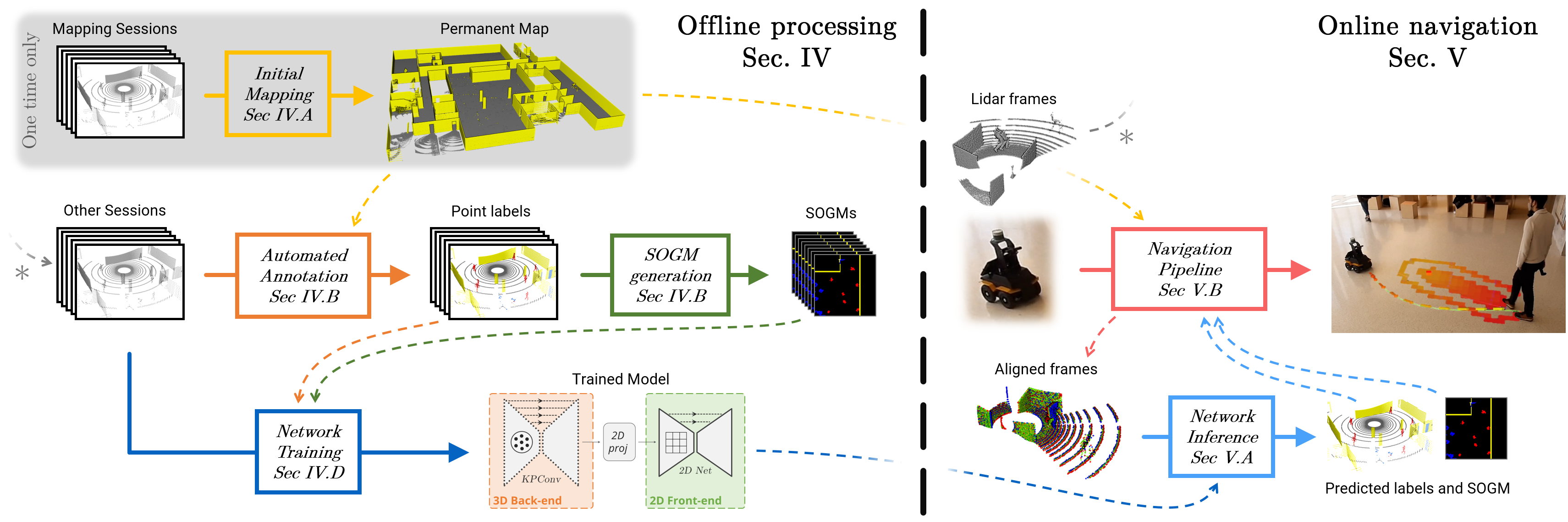}
\caption{Our approach aims at robots navigating in the same environment repeatedly. Initially, we create a point cloud map of the environment. Then we alternate offline processing, where a network is trained on the collected data, and online navigation, where the robot collects data (whether it uses the predictions or not.}
\label{fig_approach}
\end{figure*}

We study this question in the context of a concrete example: a robot learning on its own to navigate among humans or dynamic objects in an indoor space. Our approach allows the robot to predict the location of obstacles in a short future horizon (a few seconds), and plan its way around them. A deep neural network predicts these locations as Spatiotemporal Occupancy Grid Maps (SOGMs), which contain occupancy probabilities in space an time, as shown in Figure \ref{fig_intro}. We use Self-Supervised Learning, which means the training data and annotations are collected automatically. After the robot navigated in a dynamic scene, our annotation pipeline can label 3D lidar points with semantic information and generate past SOGMs, without any human annotation. We supervise the training of our network with this annotated data. Then the robot can navigate with our network prediction integrated in the navigation system, and thus anticipate the movements of dynamic obstacles.
In this paper, we provide a detailed description of the collection of algorithms required for these various tasks, for a complete view of the overall approach, as illustrated in Figure \ref{fig_approach}.

Some of the algorithms we use have already been introduced in two of our previous works. In the first one \cite{thomas2021self}, we described how to automatically annotate 3D lidar points, and train a deep network to predict these 3D labels. In the second one \cite{thomas2022learning}, our system learned to predict the future of dynamic scenes as SOGMs. Until now, we only evaluated results in a simulated environment. In this work, we build on these two previous papers and improve our approach with \textbf{three novel contributions}:

\begin{itemize}
\item{a new lidar and SOGM automated annotation pipeline working with real noisy lidar data.}
\item{a new closed-loop navigation system using our network predictions.}
\item{experiments on a real robot, with the data published as an open dataset.}
\end{itemize}

After a literature review in \textbf{Section \ref{sec_related}}, where we highlight the uniqueness of our approach, we define the building blocks used in different parts of our approach in \textbf{Section \ref{sec_algos}}. Our localization and mapping method PointMap, our main annotation tool that estimates occupancy probabilities PointRay, and point-cloud mathematical morphology operators to help reduce the noise in the annotations. 

We dedicate \textbf{Section \ref{sec_offline}} to the offline half of our approach. First, a 3D point-cloud map of the environment is created. Then, our lidar annotation algorithm identifies four lidar point labels: \textit{ground}, \textit{permanent} (structures such as walls), \textit{movable} (still but movable obstacles such as chairs), and \textit{dynamic} (moving objects such as people); and generates SOGMs. Finally, this labeled data is used for the training of our 3D-2D feedforward architecture, which only takes three consecutive lidar frames as input, and predicts 3D point labels and the future SOGM. \textbf{Section \ref{sec_online}} focuses on our online navigation system. The network inference is post-processed to obtain Spatiotemporal Risk Maps (SRM) and 3D point labels, that easily connect and interact with the rest of the navigation system. 

The simulation experiments, listed in \textbf{Section \ref{sec_simu}}, provide quantitative evaluations of our navigation system, as they allow the use of groundtruth, and multiple repetitions. We compare the efficiency and safety our navigation system when using different types of predictions. The experiments we conduct in the real world \textbf{Section \ref{sec_real}} are crucial to validate that our method generalizes to real applications. We study the predicted SOGMs quantitatively and qualitatively, and provide anecdotal examples of navigation on a real robot.

Our results are best viewed in the supplementary video\footnote{\urlstyle{sf}Video: \textcolor{mypink}{\url{https://huguesthomas.github.io/tro_2022_video.html}}}. In addition, we publish the data collected during our experiment as \textbf{a new dataset: UofT-Indoor-3D (UTIn3D)}\footnote{\urlstyle{sf}Dataset: \textcolor{mypink}{\url{https://github.com/utiasASRL/UTIn3D}}}. It includes the lidar frames, the localization, and the labels provided by our automated annotation approach. We believe it will be valuable to the community given the rarity of indoor 3D lidar dataset with many pedestrians. Along with the dataset, we aim to facilitate the reproduction of our results and encourage research in this direction, with detailed open-source code\footnote{\urlstyle{sf}Code: \textcolor{mypink}{\url{https://github.com/utiasASRL/Crystal_Ball_Nav}}}.

%
%
%
%
%
%
%
%
%
%
%
%

\section{Related work}
\label{sec_related}


Navigation around dynamic obstacles is well studied in robotics. In terms of predicting obstacles' future motions, \cite{ziebart2009planning, kitani2012activity} learned a distribution of possible pedestrian trajectories using inverse optimal control to recover a set of agent preferences consistent with prior demonstration. Following these preliminary works, various solutions to dynamic obstacle forecasting have been explored, that we study in this literature review.

Our work is unique compared to other deep-learning-based approaches for navigation in dynamic scenes. It stands out because of three crucial properties:

\begin{itemize}
\item{\textbf{Self-Supervised}: We use annotation generated automatically and not by humans.}
\item{\textbf{End to end}: Our predictions do not rely on other algorithms such as object detection and tracking.}
\item{\textbf{Pointwise/non-parametric}: the computation cost of our method does not increase with the number of agents in the scene.}
\end{itemize}

%
%
%
%
%
%

\subsection{Mapping, Localization, and Occupancy Probabilities}

ICP-based simultaneous localization and mapping (SLAM) algorithms are widely used in robotics, with many variants \cite{pomerleau2013comparing, zhang2014loam, mendes2016icp, deschaud2018imls}. We designed our PointMap SLAM with a focus on simplicity and efficiency. Similarly to \cite{deschaud2018imls}, we keep a point cloud with normals as the map, but we update the normals directly on the lidar frames with spherical coordinate neighborhoods. Our approach targets the problem of robots that navigate in the same environment repeatedly. Therefore we prefer to use PointMap as a mapping tool first and then only rely on the frame-to-map alignment for localization.

Computing occupancy probabilities with ray-casting is also a common technique in the literature. Used at first for 2D occupancy grid mapping \cite{moravec1985high}, it was later adapted for 3D mapping \cite{izadi2011kinectfusion, hornung2013octomap}. In our case, PointRay computes occupancy probabilities on a point cloud instead of a grid, similarly to \cite{pomerleau2014long}, and therefore only models free space where points have been measured. Our main addition is the notion of dynamic and movable labels, which we get by combining multiple sessions. Other minor differences with \cite{pomerleau2014long} include a simplification of the probability update rules and the use of frustums instead of cones around lidar rays.

Combining occupancy probabilities and real-time localization is still relatively under-explored. \cite{biswas2014episodic} propose to detect short-term and long-term movables similar to our dynamic and movable labels. However, they compute their short-term and long-term features with ray-tracing in a 2D map, while we propose to train a deep network able to predict them directly in the 3D lidar points based on the appearance of the point clouds. Predicting movable points with deep networks was also suggested by \cite{dewan2017deep}. However, they chose a 2D architecture, FastNet, using lidar depth images, and they only predict an objectness score from a human-annotated training set.

%
%
%
%
%
%

\subsection{Self-supervised Learning for Robotics Application}

Self-supervised learning is a form of unsupervised learning where the data provides supervision. In robotics, this term usually refers to methods using an automated acquisition of training data \cite{sofman2006improving, lookingbill2007reverse, hadsell2009learning, brooks2012self, ridge2015self, nava2019learning}. These approaches often exploit multiple sensors during the robot's operation. In our case, we only use a 3D lidar and an algorithm that provides automated annotation for the data. Deep-learning-based semantic SLAM algorithms have been proposed, using either camera images \cite{zhang2018semantic, wang2019unified}, lidar depth images \cite{chen2019suma++}, or lidar point clouds \cite{sun2018recurrent}, but they always rely on human-annotated datasets, whereas our method learns on its own. To the best of our knowledge, our approach is the first to use multi-session SLAM and ray-tracing as annotation tools for the training of semantic segmentation networks.

%
%
%
%
%
%

\subsection{Object Tracking and Trajectory Prediction}

Following the success of recurrent neural networks (RNNs) and in particular long short-term memory networks (LSTMs) \cite{alahi2016social, gupta2018social}, the idea of trajectory prediction has received a lot of attention. It requires the obstacles to be isolated as distinct object and tracked. \cite{katyal2020intent} uses an LSTM-based network to predict obstacle trajectories and plan around them, and \cite{peddi2020data} exploits a Hidden Markov Model to predict future states from a history of observations. Similarly, \cite{sathyamoorthy2020frozone} detects individual obstacles and predicts their speeds to avoid ``freezing zones''. Similar object-centric methods are also used in the context of autonomous driving \cite{luo2018fast, casas2018intentnet}. However, they all rely on detection and tracking predictions and do not easily incorporate multi-modal predictions, problems we do not face with our point-centric approach. Closer to our work, \cite{jain2020discrete} predicts future human motions as 2D heat maps, implicitly handling multi-modality, but  still relies on object-level predictions and is also limited to 2D inputs, where our method leverages 3D features that are more descriptive.

A fair comparison between trajectory prediction methods and our work is nearly impossible because they rely on detection and tracking algorithms, and are usually evaluated with respect to each object in the scene. On the contrary, our occupancy grid predictions are different in nature and are evaluated as a representation of the whole scene. We argue that the difficulty that object-centric methods have to scale with a high number of agents and handle multi-modality, which are two things inherently handled by our approach, justify the relevance of our contributions. In addition labelling object automatically is harder than labelling points automatically.

%
%
%
%
%
%

\subsection{Reinforcement Learning for Navigation}

Reinforcement Learning has been used extensively in recent years to replace standard motion planning algorithms \cite{chen2017socially, long2018towards, liang2020crowdsteer, sathyamoorthy2020densecavoid, everett2021collision, strudel2020learning, liu2020robot}. However, standard local planners have proven to be very reliable methods, especially when it comes to producing dynamically feasible trajectories, which most RL methods fail to do. Even when the feasibility is ensured \cite{patel2021dwa}, the whole planning algorithm is embedded into a black box end-to-end neural network, which is difficult to tune, debug, and interpret. We chose to keep a standard local planner, with its guarantees and interpretability, and use a self-supervised deep learning method to predict the future motion of dynamic obstacles.

%
%
%
%
%
%

\subsection{Occupancy Grid Maps Predictions}

OGM prediction approaches are the closest to our work and can be separated into two groups either using handcrafted or learned features. Handcrafted approaches usually rely on a model for human motion prediction. \cite{pierson2019dynamic} predicts a Dynamic Risk Density based on the occupancy density and velocity field of the environment. \cite{huang2020safe} extends this idea with a Gaussian Process regulated risk map of tracked pedestrians and cars.  Other recent works focus on adapting the uncertainty of the predictive model \cite{fisac2018probabilistically, bajcsy2019scalable, bansal2020hamilton}, using real-time Bayesian frameworks in closed-loop on real data. These methods either rely on object detection and tracking or are based on instantaneous velocities, and are not able to predict future locations of obstacles accurately. 
Learned methods, usually based on video frame prediction architectures \cite{lotter2016deep, wang2018predrnn++, wang2019memory}, are better at predicting complex futures. \cite{mohajerin2019multi} introduces a difference-learning-based recurrent architecture to extract and incorporate motion information for OGM prediction. \cite{schreiber2020motion} presents an LSTM-based encoder-decoder framework to predict the future environment represented as Dynamic Occupancy Grid Maps. They introduce recurrent skip connections into the network to reduce prediction blurriness. To account for the spatiotemporal evolution of the occupancy state of both static objects and dynamic objects, \cite{toyungyernsub2020double} proposes a double-prong network architecture.

However, two major differences remain in our approach. First, these methods all take previous OGMs as input, effectively losing valuable information in the shape patterns that common 3D sensors can capture. To our knowledge, we are the first to fill this gap in the literature, by incorporating 3D features in the forecasting of OGMs with the 3D backbone of our network. Second, we are also the first, to our knowledge, to predict a sequence of OGMs without recurrent layers. We argue feedforward architectures are easier to train and understand. Eventually, our network can make a distinction between different semantic classes, leveraging interactions between them, when predicting future occupancy.

%
%
%
%
%
%
%
%
%
%
%
%

\section{Algorithms Used in Our Pipeline}
\label{sec_algos}

Before presenting our approach as a whole, this section introduces the key algorithms that are used throughout our pipeline.

%
%
%
%
%
%

\subsection{PointMap: ICP-based 3D Localization and Mapping}

\begin{table}[b]
\caption{ICP Configuration for PointMap with a Velodyne HDL-32E.}
\setlength\tabcolsep{0.5pt}
\begin{footnotesize}
\begin{center}
\begin{tabular}{ L{1.6cm}  L{6.8cm}   }
\Xhline{2\arrayrulewidth}
\textbf{Elements} & \textbf{Choices} \TBstrut\\
\Xhline{2\arrayrulewidth}
Filters & Subsample frame with a 12cm grid. \TBstrut\\
\hline
Matcher & Sample 600 points at each iteration according to $w_\mathrm{icp}$.\Tstrut\\
 & Match with single nearest neighbor (or ground plane). \TBstrut\\
\hline
Rejection & If pt2pt distance $> 2$m. \Tstrut\\
 & If pt2pl distance $> 12$cm (except for the first iteration). \Bstrut\\
\hline
Distance & Optimize point-to-plane distance. \TBstrut\\
\hline
Convergence & Stop at a maximum of 100 iterations. \Tstrut\\
 & Stop if relative motion goes below 0.01m and 0.001rad. \Bstrut\\
\Xhline{2\arrayrulewidth}
\end{tabular}
\end{center}
\end{footnotesize}
\label{Table_icp} 
\vspace{-3ex}
\end{table}

PointMap \cite{thomas2021self} is our SLAM algorithm, which has two components: an Iterative Closest Point (ICP) localization solution that aligns lidar frames on a point cloud map, and a mapping function that updates the map with the aligned frame. Each function can be used independently or together in a SLAM mode.

For a detailed description of the ICP algorithm, we refer to a previous in-depth review \cite{pomerleau2013comparing}. Following their work, we use the same elements to characterize our ICP: the data filters, the matcher, the outlier rejection, the distance function, and the convergence tests. Our choices for each element are listed in Table \ref{Table_icp}. Most of these are based on previous works \cite{mendes2016icp, deschaud2018imls, thomas2021self}, taking into account that we use a Velodyne HDL-32E sensor. Some elements, including matching, are simplified for efficiency. We use the latest odometry of the robot as the initial pose to solve the initialization issue common to most ICP solutions. In the following, we define transformations by their rotation and translation components $(\mathbf{R}, \mathbf{t})$. In comparison to \cite{thomas2021self}, we handle the  motion-distortion effect of real data within the iterative process of ICP. At each ICP iteration, after estimating the transformation $(\mathbf{R}_1, \mathbf{t}_1)$ at time $t_1$ (last timestamp of the current lidar frame), and before matching neighbors, we apply motion distortion. We have the poses $(\mathbf{R}_0, \mathbf{t}_0)$ and $(\mathbf{R}_1, \mathbf{t}_1)$ of the lidar at time $t_0$ and time $t_1$, therefore for any point stamped with a time $t \in \left[t_0, t_1 \right]$, its pose $(\mathbf{R}, \mathbf{t})$ is computed as

\begin{equation}
\label{eq1}
    \left.
    \begin{array}{ll}
    \omega = \left(t - t_0\right) / \left(t_1 - t_0\right) \: ,\\
    \mathbf{t} = \omega \mathbf{t}_1 + \left(1 - \omega\right) \mathbf{t}_0 \: ,\\
    \mathbf{R} = \mathrm{Slerp}\left(\mathbf{R}_0, \mathbf{R}_1, \omega \right) \: ,
    \end{array}
    \right.
\end{equation}

\noindent where $\mathrm{Slerp}$ is the spherical linear interpolation \cite{shoemake1985animating}. Note that during ICP convergence, we chose $t_0$ as the beginning of the previous frame instead of the end of the previous frame. Otherwise, the smallest mistake made when estimating the previous pose will cause issues and sometimes divergence. 

In addition, we use an optional ground plane heuristic in the optimization. This heuristic, particularly useful for indoor scenarios, assumes that the ground is a planar horizontal surface of height $z=0$. During ICP optimization, any point considered ground can be matched to this plane instead of its nearest neighbor.

The second component of PointMap, the map update function, adds the information from an aligned frame to the map. We use the same update function described in \cite{thomas2021self}, where the map is defined as a sparse voxel grid. The voxel size is ${dl}_{\mathrm{map}} = 3$cm and we keep only one point per voxel, with its normal and its score. When using an initial map for localization, we can create a secondary map that we ignore for localization, but keep for further processing, which is particularly useful for the buffer creation step shown in Figure \ref{fig_annot} (a).

%
%
%
%
%
%

\subsection{PointRay: Ray-tracing Occupancy Probability}

We use PointRay \cite{thomas2021self} to compute occupancy probability in a point cloud map. The occupancy probability of each point in a map can be found thanks to the data provided by lidar frames. Indeed, each frame provides two kinds of information: occupied space where the points are located, and free space along the lidar rays.  If a location exists in the map but gets passed through by multiple rays, it will have a low occupancy probability. We follow the idea from \cite{pomerleau2014long} and use the projection of the map in the frame spherical coordinates to model the lidar rays. The occupancy probabilities are deduced from the distance gap between frame points and map points. 

PointRay assigns two values to each point of the map $x_i$, in a voxel $i$: $n_i$ the number of times this voxel has been seen, and $o_i$ the number of times this voxel was occupied. For each lidar frame in the list, PointRay first gets the list of occupied voxels, and increments both $n_i$ and $o_i$ for them. For the rest of the points, PointRay verifies if they are seen by a free space frustum in the spherical coordinates $(\rho, \theta, \phi)$. The frustums are defined as the pixels of a 2D grid in the $\theta$ and $\phi$ spherical dimensions. Each pixel (or frustum) stores the smallest point distances to the lidar origin ($\rho$ spherical coordinate). The resolution of the grid is $d\theta = 0.33^\circ$ and $d\phi = 0.5^\circ$, but in the $\theta$ dimension, the lidar resolution is variable, so we use nearest-neighbor interpolation to fill the empty pixels along this dimension. 

The verification is done by projecting the map in the same frustum grid. To reduce the effect of motion distortion, we cut the lidar frame in $n_\mathrm{slices}=16$ slices along the azimuth and perform the map projection with the median pose of each slice. Given the small slicing angle, the distortion effect is negligible. Although it is slower to compute the occupancy probability one slice at a time, we alleviate the computational cost by only projecting map points that are in the slice area. For every projected map point $x_i$, PointRay increments $n_i$ (and not $o_i$) only if the two following conditions are respected:

\begin{equation}
\label{eq2}
    \left.
    \begin{array}{ll}
    \mathrm{cond}_A: &\rho_i < \rho_0 - \mathrm{margin}(\rho_0) \: , \\
    \mathrm{cond}_B: & \left|n_z\right| > \cos(\beta_{\mathrm{min}}) \quad \mathrm{OR} \quad \alpha < \alpha_{\mathrm{max}} \: , \\
    \:
    \end{array}
    \right.
\end{equation}

\noindent where $\rho_0$ and $\rho_i$ are the frustum radius and point radius respectively, $\mathrm{margin}(\rho_0) = \rho_0 \max(d\theta, d\phi) / 2$ is the largest half size of the frustum at this particular range, $n_z$ is the vertical component of the point normal, and $\alpha$ the incidence angle of the lidar ray with this normal. $\alpha_{\mathrm{max}} = \frac{5\pi}{12}$ and $\beta_{\mathrm{min}} = \frac{\pi}{3}$ are heuristic thresholds. $\mathrm{cond}_A$ ensures that, at any range, a planar surface whose incidence angle is less than 45$^{\circ}$, will not be updated as free space. $\mathrm{cond}_B$ handles the wider incidence angles, we do not want to update for extreme incidence values, except if the normal is vertical because tables are usually nearly parallel to the lidar rays and would never be updated otherwise. Because of $\mathrm{cond}_B$, ground points are more likely to have low occupancy probabilities, but we take care of that by extracting the ground as a distinct semantic class, unaffected by ray-tracing.

When all the lidar frames of a session have been processed, the final occupancy probability for each point $x_i$ of the map is computed as $p_i = o_i / n_i$. A point has to be seen at least $n_\mathrm{min} = 10$ times to be considered valid, otherwise, $p_i = 0.5$.

%
%
%
%
%
%

%
\subsection{PointMorpho: Pointcloud Mathematical Morphology}

In our annotation pipeline, one of the issue we face when dealing with real data is the noise. As shown in the middle-bottom picture of Figure \ref{fig_annot}, the point labels found automatically can be noisy. In this section, we define Mathematical Morphology operators for point clouds to help reduce the noise in the annotations, through spatial smoothing.

Mathematical Morphology regroups techniques for processing geometrical structures and was originally developed for binary images, with four basic morphological operators: erosion, dilation, opening, and closing \cite{matheron2002birth}. Some works have tried to adapt mathematical morphology to point clouds, by considering points as positive elements, and empty space as negative elements \cite{calderon2014point, balado2020mathematical}.

\begin{figure}[t]
    \centering
    \includegraphics[width=0.999\columnwidth, keepaspectratio=true]{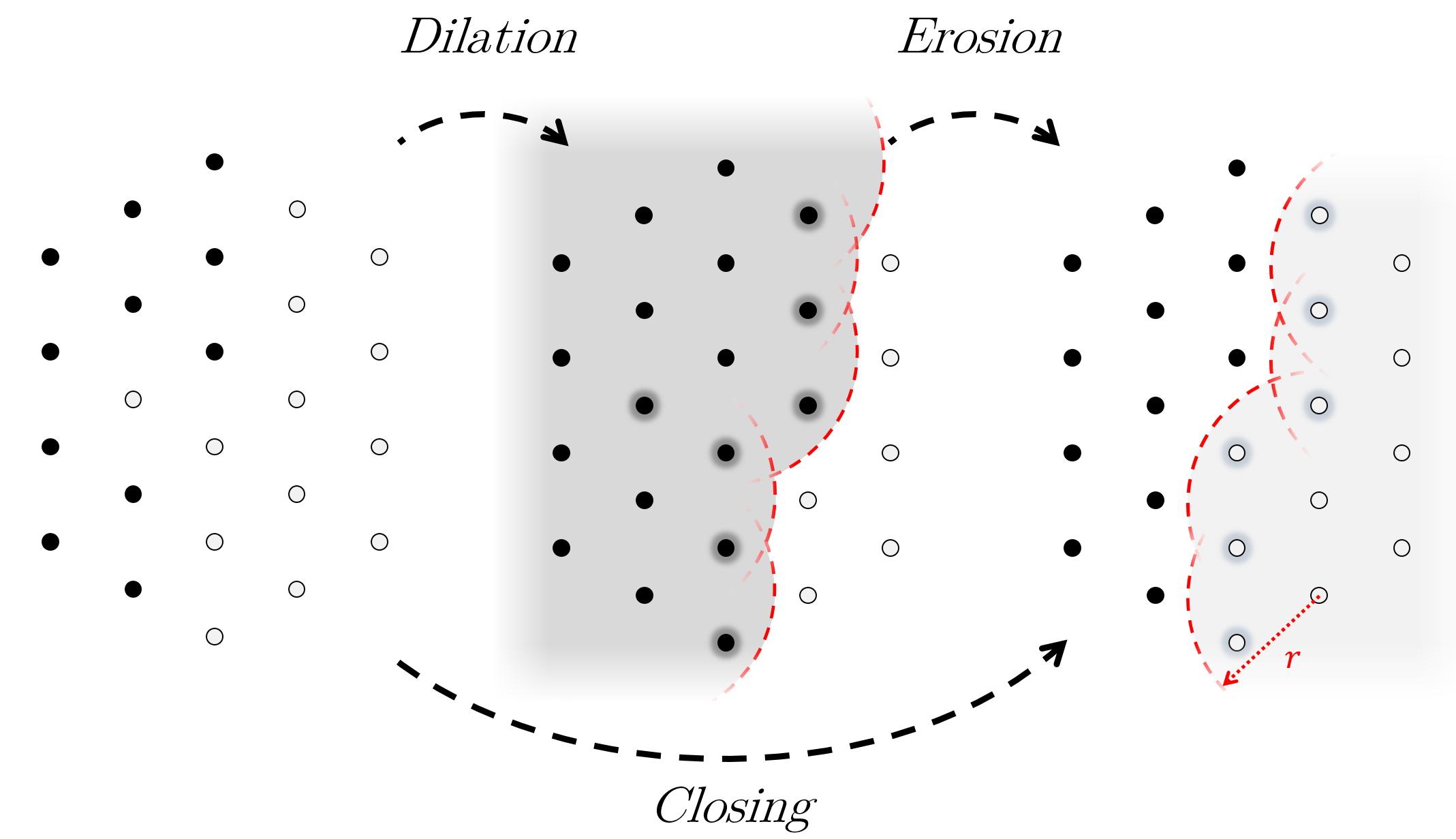}
    \caption{Illustration of point morphology operations with a radius $r$.}
    \label{fig_morpho}
\end{figure}

In this work, we are interested in a simpler problem: applying mathematical morphology on point clouds where some points are considered positives and the other negatives (see Figure \ref{fig_morpho}). We only consider a sub-problem where the structural element is a sphere of radius $r$, the positive elements are denoted as point cloud $A$ and the negative elements as point cloud $B$. Therefore, the morphology operations can be described simply:
\begin{itemize}
\item{\textbf{Dilation}: $\mathcal{D}_r\left(A, B\right) = A \cup \mathcal{N}_{B, r}\left(A\right) $}
\item{\textbf{Erosion}: $\mathcal{E}_r\left(A, B\right) = A \setminus \mathcal{N}_{B, r}\left(A\right) $}
\item{\textbf{Closing}: $\mathcal{C}_r\left(A, B\right) = \mathcal{E}_r\left( \mathcal{D}_r\left(A, B\right), \mathcal{E}_r\left(B, A\right)  \right)$}
\item{\textbf{Opening}: $\mathcal{O}_r\left(A, B\right) = \mathcal{D}_r\left( \mathcal{E}_r\left(A, B\right), \mathcal{D}_r\left(B, A\right)  \right)$}

\end{itemize}

Where $\mathcal{N}_{B, r}\left(A\right) = \left\{ p \in A \; | \; \exists q \in B: \left\|p - q\right\| \leq r \right\}$ is the subset of $A$ in the neighborhood of $B$. As we will see in Section \ref{sec_annot}, these tools are particularly efficient for removing noise and undesired artifacts in the annotations.

%
%
%
%
%
%
%
%
%
%
%

\section{Offline processing}
\label{sec_offline}

%
%
%
%
%
%

\subsection{Initial Mapping}

In this work, we assume that our robot will be navigating in the same space for some time, and thus will use a map of this environment for localization. This map will also be used during the annotation process. As explained in our previous works, for the annotation, we need the map to contain only \textit{ground} and \textit{permanent} points. It is also convenient for localization, as these points are usually from large planar surfaces, easy to localize against. 

The mapping process starts by using PointMap in SLAM mode to get a first map of the environment. We then perform a loop closure using the Open3D library \cite{zhou2018open3d}. The map created from the loop-closed poses is then cleaned from its dynamic objects with PointRay. As the spaces for our real experiments have windows, we hard-code map limits, to remove the outdoor space that is mixed with reflections of indoor space. To ensure the map is also cleaned from any movable object, we perform a few other runs where the movable objects in the space have been moved to different places. These refinement runs are given to PointRay to compute occupancy probabilities. Points with a low occupancy probability are removed, and we eventually obtain a good quality map, containing only \textit{ground} and \textit{permanent} points. As it is usually the case in indoor environments, the ground is flat, allowing us to use the flat ground heuristic from PointMap. As a consequence, the ground point can easily be extracted as the horizontal plane at $z=0$.

A human could intervene at this step to move furniture around between mapping sessions, but it is not required. We assume furniture would be moved at some point in the robot life, and only do it ourselves for convenience in our experiments. We also note an interesting idea: this strategy could be used for cheap and fast 3D point cloud annotation of any object. For example, only remove the chairs first to annotate them, then remove tables, etc.

%
%
%
%
%
%

\subsection{Automated Lidar Point Annotation and SOGM Generation}
\label{sec_annot}

\begin{figure*}[t]
    \centering
    \includegraphics[width=0.999\textwidth, keepaspectratio=true]{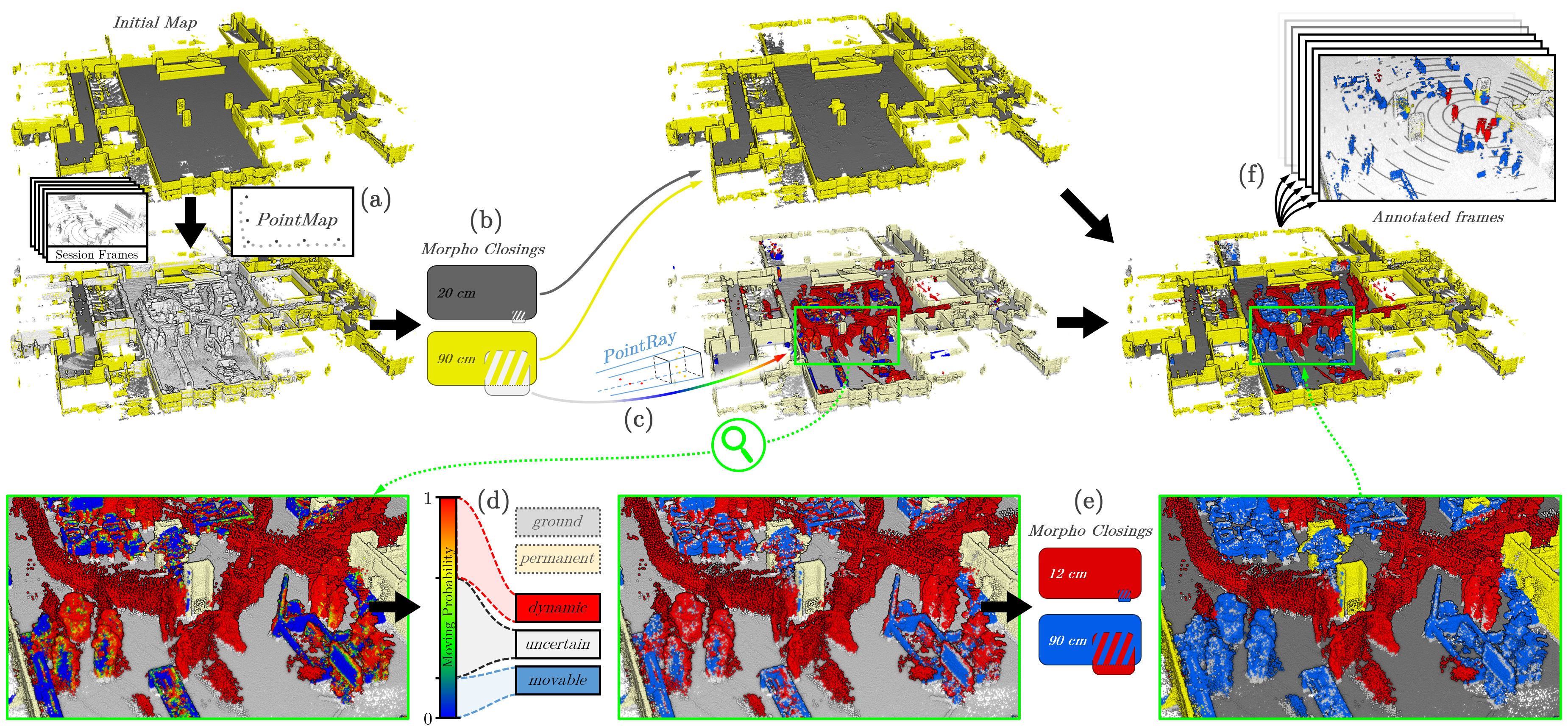}
    \caption{Illustration of our automated annotation process for real lidar frames. A buffer cloud is created with PointMap (a). The \textit{permanent} and \textit{ground} points are found with morphological closings (b), and the remaining points are annotated as \textit{dynamic} or \textit{movable} by PointRay (c-d). Noise is reduced with morphological closings (e), and the labels are projected back to the frames (f).}
    \label{fig_annot}
\end{figure*}

We use an automated annotation process to label the 3D lidar points \cite{thomas2021self} and generate training SOGMs \cite{thomas2022learning}, allowing self-supervised learning. Our network can thus learn from new situations encountered throughout the robot's life.

We first annotate lidar frames, with the combination of PointMap and PointRay, as shown in Figure \ref{fig_annot}. As opposed to \cite{thomas2021self, thomas2022learning}, we are able to handle noisy and imperfect real data. We consider that our map has already been refined and we focus on the automated annotation of the lidar point cloud labels. The lidar frames of the session are aligned on the map with PointMap, and a buffer of new points is created. To annotate the \textit{permanent} points, we perform a point-morphology closing of the map \textit{permanent} points on the buffer points ($r = 0.9$m). To annotate the ground point we perform another closing of the map \textit{ground} points on the buffer points ($r = 0.2$m). This leaves us with the remaining buffer points that are processed by PointRay, to get their occupancy probabilities, $p_i$, during this session. Points with $p_i < \tau_\mathrm{s} = 0.3$ are labeled as \textit{dynamic}, points with $p_i > \tau_\mathrm{m} = 0.6$ are labeled as \textit{movable}, and the remaining points are left \textit{uncertain}. We conduct a noise removal step consisting of a first closing of \textit{dynamics} on \textit{movables} ($r = 0.12$m) to clean the isolated or tiny groups of \textit{movable} points inside dynamic areas. And then a second larger closing of \textit{movables} on \textit{dynamics} ($r = 0.9$m) to ensure we only keep large groups of dynamic points. Eventually, the annotations are projected back on the lidar frames, taking into account motion distortion.

\begin{figure}[!b]
    \centering
    \includegraphics[width=0.999\columnwidth, keepaspectratio=true]{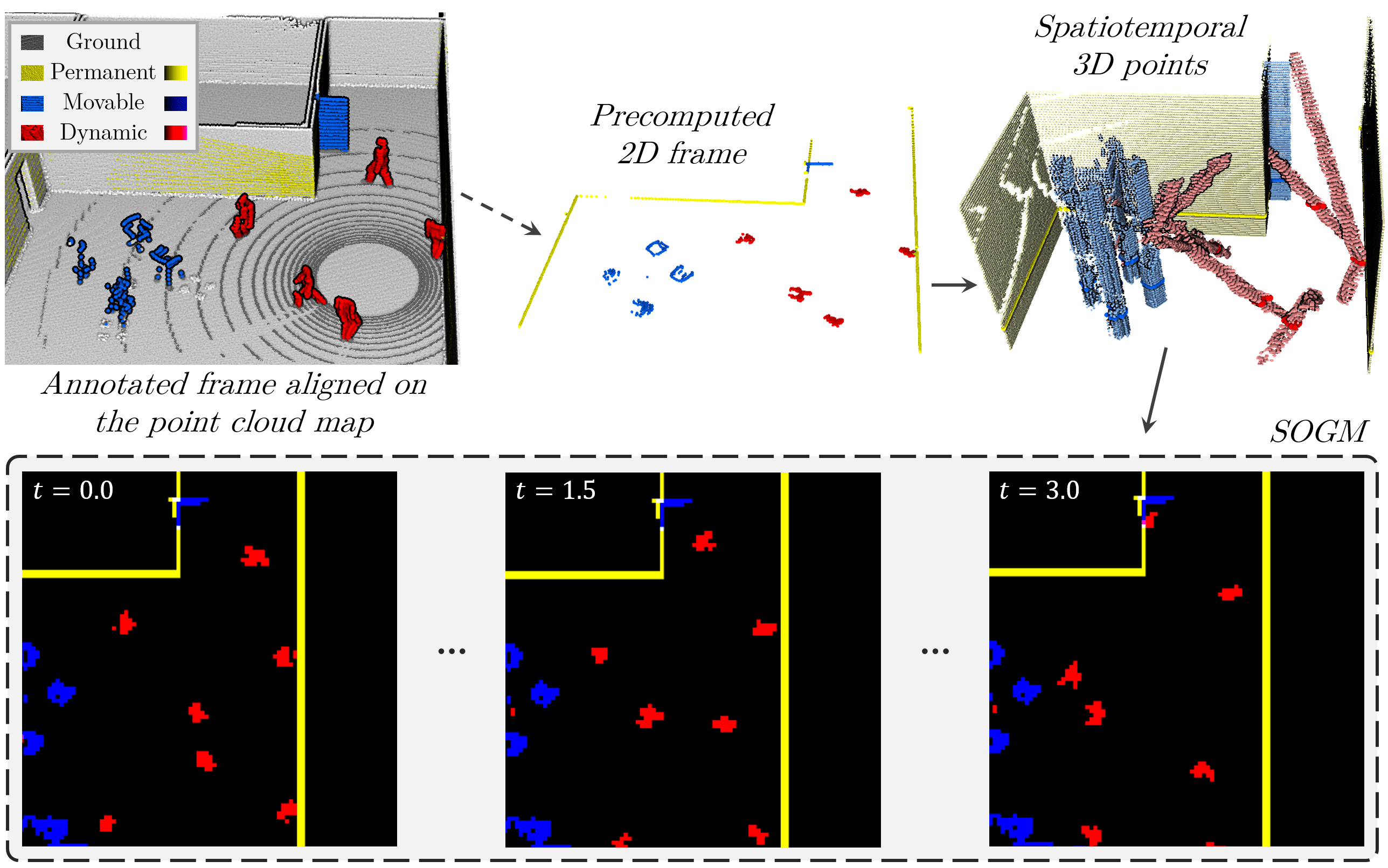}
    \caption{During pre-processing, every frame is semantically filtered and projected in 2D. During training, the 2D frames are stacked in 3D according to their timestamps and projected to a 3D grid to create the SOGMs.}
    \label{fig_sogm}
\end{figure}

Now that we have annotated lidar frames, we can generate SOGMs automatically. Our SOGMs have three channels, one for each obstacle class: \textit{permanent}, \textit{movable}, and \textit{dynamic}. Because we want to be able to augment the training data with rotations, it is better to save our 2D labels as intermediate 2D point cloud structures that can easily be rotated and then transformed into SOGMs during training. An annotated 2D point cloud is computed and saved for each lidar frame by removing non-obstacle points, projecting the remaining points on a horizontal plane, and subsampling them with a grid size of $3$cm. To reduce the noise, we remove isolated dynamic points from these precomputed 2D point clouds, and perform a point morphology opening of the \textit{dynamic} points by the static points (\textit{permanent} + \textit{movable}), with $r = 0.3$m.

At training time, we stack the 2D points in a third dimension according to their timestamps, apply data augmentation, and project them to a SOGM structure of spatial resolution $dl_\mathrm{2D}=12$cm and temporal resolution $dt=0.1$s. The \textit{permanent} and \textit{movable} occupancies from all time steps of the SOGM are merged because they are not moving. Therefore, in addition to the future locations of dynamic obstacles, our network also learns to complete partially seen static objects.

%
%
%
%
%
%

\subsection{Network Architecture for Lidar Segmentation and SOGM Prediction}

\begin{figure}[t]
    \centering
    \includegraphics[width=0.999\columnwidth, keepaspectratio=true]{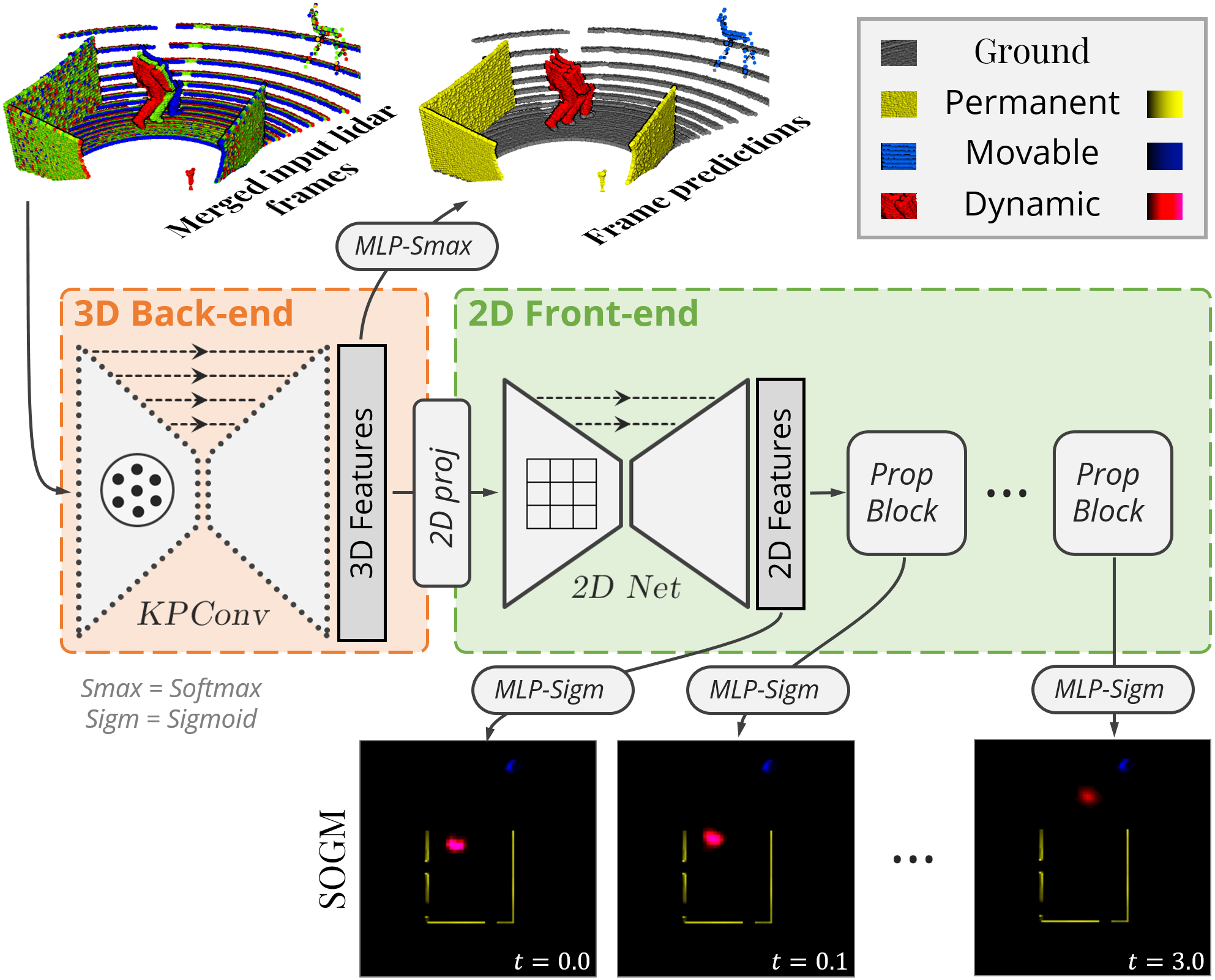}
    \caption{Illustration of our 3D-2D feedforward architecture. The 3D back-end is a 5-layer KPConv architecture, producing features at the point level. The 2D front-end is composed of a 3-layer 2D U-Net architecture, followed by consecutive convolutional propagation blocks.}
    \label{fig_net}
\end{figure}

Our network architecture (Figure \ref{fig_net}) is composed of two parts, a 3D back-end, and a 2D front-end. The 3D back-end is a KPConv network \cite{thomas2019kpconv} predicting a semantic label for each input point. Predicting 3D labels helps the network training by providing an additional supervisory signal and ensures that rich features are passed to the 2D front-end. We keep the KP-FCNN architecture and parameters of the original paper: a U-Net with five levels, each composed of two ResNet layers, refer to \cite{thomas2019kpconv} for details. The network input is a point cloud made from $n_\mathrm{f}=3$ lidar frames aligned in the map coordinates and merged. We only keep the points inside a $R_\mathrm{in}=8$m radius, as we are interested in the local vicinity of the robot. Each point is assigned a one-hot $n_\mathrm{f}$-dimensional feature vector, encoding the lidar frame to which it belongs. To help with computational speed for inference on a real robot, the input point density is controlled using a relatively large grid subsampling size ($dl_\mathrm{3D}=12$cm). It is equal to the PointMap subsampling size, allowing us to reuse the subsampled cloud aligned and rectified by PointMap as is.

The 3D point features of dimension are passed to the 2D front-end with a grid projection using the same spatial resolution $dl_\mathrm{2D}$ as the SOGM. The size of the grid is determined as the inscribed square in the $R_\mathrm{in}$-radius circle: $h_\mathrm{grid} = w_\mathrm{grid} = 94$. Features from points located in the same cell are averaged to get the new cell features. The features of the empty cells are set to zero. The obtained 2D feature map of dim is then processed by an image U-Net architecture with three levels, each composed of two ResNet layers, to diffuse the information contained in sparse locations to the whole grid. This dense feature map is used to predict the initial time step of the SOGM. Then, it is processed by successive propagation blocks, each composed of two ResNet layers. The output of each propagation block is used to predict the corresponding time step of the SOGM. We define the final prediction time $T=4.0$s, meaning that our SOGMs have  $n_T = T / dt + 1 = 41$ time steps in total. More details and hyperparameters can be found in our implementation. Note that the \textit{permanent} and \textit{movable} predictions are redundant but we keep them to help the network to keep the knowledge of their location, and to learn better interactions between the classes further into the future.

%
%
%
%
%
%

\subsection{Network Training}

The training loss of our network is a combination of two loss functions. The standard semantic segmentation loss of a KPConv network $L^\mathrm{3D}$, and a loss function applied to each SOGM prediction layer $L^\mathsf{2D}_k$. We define it as

\begin{equation}
\label{eq2}
    L_\mathrm{tot} = \lambda_1 L^\mathrm{3D} + \lambda_2 \sum\limits_{k<n_T}{  \frac{L^\mathrm{2D}_k}{n_T} \: , }
\end{equation}

\noindent where $\lambda_1=1.0$, $\lambda_2=10.0$. $L^\mathrm{3D}$ is the standard cross entropy loss used in the original KPConv network, and $L^\mathsf{2D}_k$ is a Binary Cross-Entropy loss applied to layer $k$ of our SOGM predictions:

\begin{equation}
\label{eq3}
    L^\mathrm{2D}_k = \sum\limits_{i \in M_k}{{BCE}(x_{k,i}, y_{k,i})}  \: ,
\end{equation}


\begin{figure}[b]
    \centering
    \includegraphics[width=0.999\columnwidth, keepaspectratio=true]{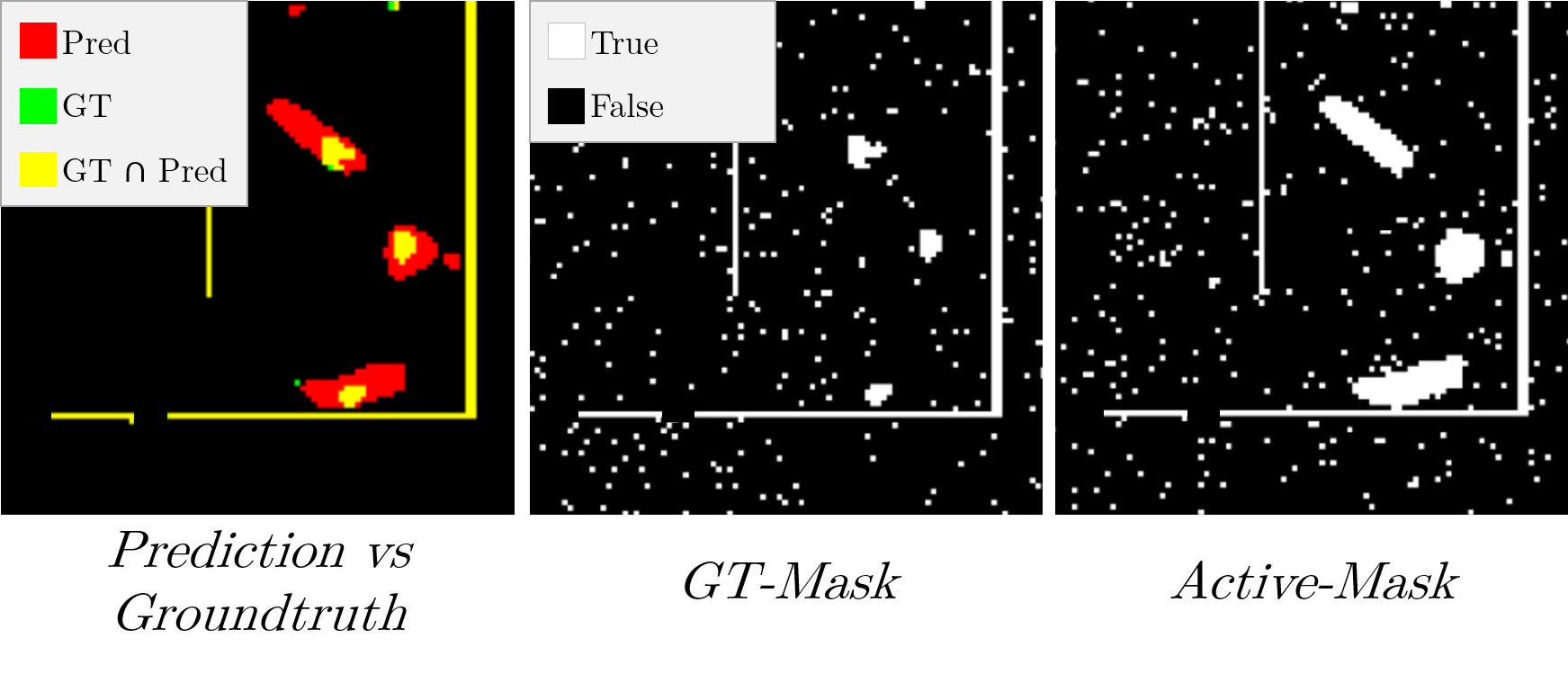}
    \caption{Loss masks reducing the influence of empty pixels during training.}
    \label{fig_mask}
\end{figure}

\noindent where $x_{k,i}$ is the network logit at the pixel $i$ of the time-step layer $k$ in the SOGM, $y_{k,i}$ is its corresponding label and $BCE$ stands for Binary Cross-Entropy. Note that for clarity, we use a simple index $i$ for 2D pixels. The SOGM loss is thus a masked Binary Cross-Entropy, where the mask $M_k$ is here to help ignore the over-represented empty pixels and focus on the positive examples. We first tried a mask covering the positive label values in addition to some random pixels (GT-Mask), but then improved it to cover the union of positive labels and positive prediction pixels (Active-Mask) to help reduce the false positives (see Figure \ref{fig_mask}).

Our network is trained with PyTorch, with an SGD optimizer. The initial learning rate is $0.01$, and decayed by 0.96 at every epoch (equivalent to 0.1 every 60 epochs). In this setup, our input point clouds contain on average $20$k points. We use a variable batch size targeting $B=6$, for an average of $85$k points per batch. During training, we only use rotation augmentation around the vertical axis. To cope with unbalanced classes in real data, we implement a sampling strategy targeting input examples containing \textit{dynamic} points. Instead of sampling frames randomly during training, we chose the frames that contain \textit{dynamic} points more often than the rest of the frames (with a ratio of 10:1). We also have the possibility to train a network with a combination of real and simulation examples (with a customizable ratio), a strategy that we evaluate in section \ref{sec_real_preds}. The rest of the training parameters are kept identical as in the original KPConv paper \cite{thomas2019kpconv}, and more details can be found in our open-source implementation.

%
%
%
%
%
%
%
%
%
%
%
%

\section{Online Navigation}
\label{sec_online}

%
%
%
%
%
%

\subsection{Network Inference and Post-processing}

During navigation, our network receives lidar point clouds sent by PointMap. They are already subsampled to $dl_\mathrm{3D}=12$cm, aligned on the map, and motion rectified. When three consecutive point clouds have been received, the CPU processing kicks in and performs all the necessary operations including merging frames, subsampling layer points, computing neighbors, etc. As soon as the CPU pre-processing is over, the GPU computes a forward pass of the network and gets the predicted SOGM, which contains the future occupancy locations up to 4 seconds after the input lidar timestamp.

\begin{figure}[b]
    \centering
    \includegraphics[width=0.999\columnwidth, keepaspectratio=true]{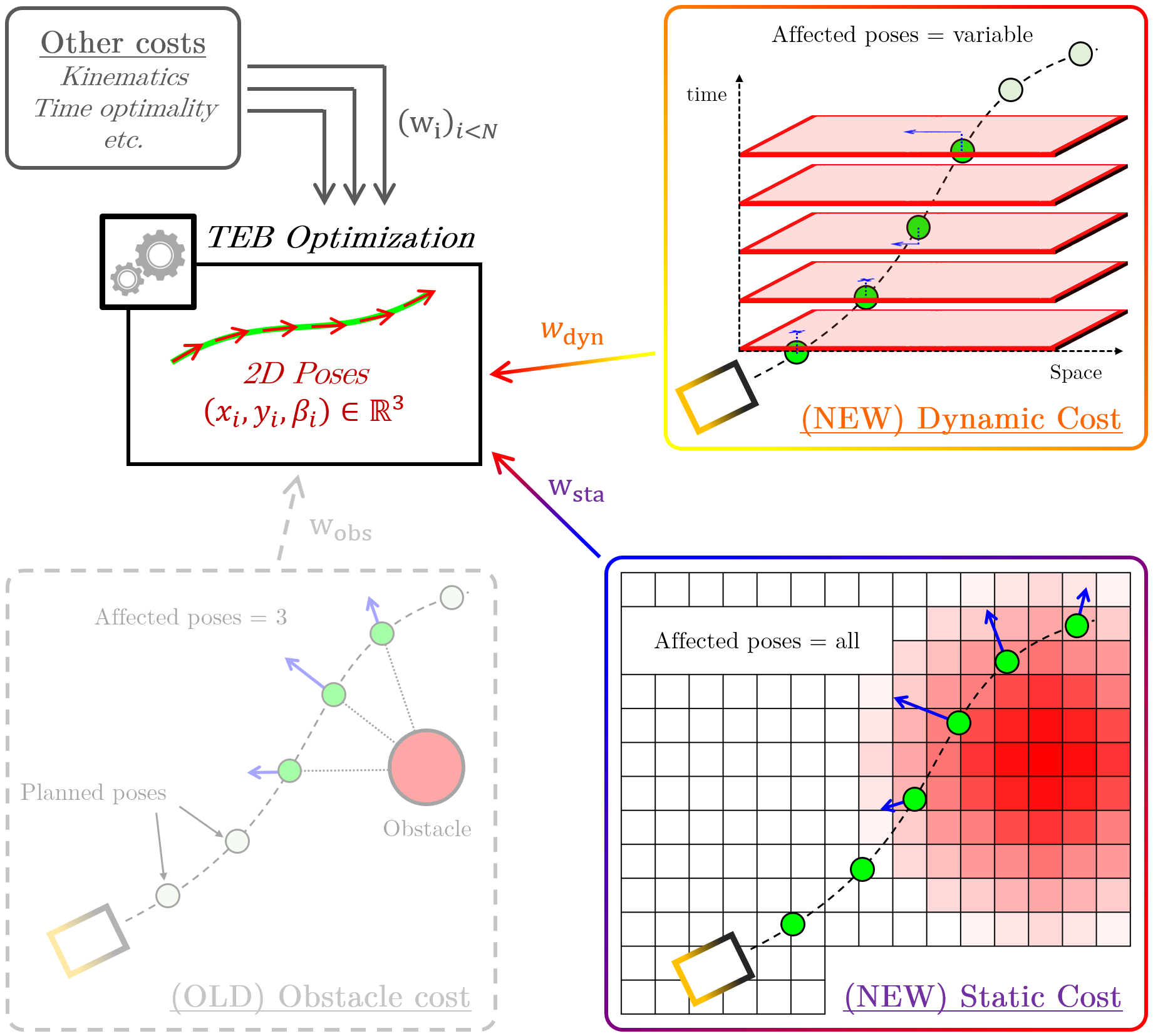}
    \caption{Illustration of TEB optimization costs. To replace the obstacle cost, we define a static cost and a dynamic cost. Our new costs use risk maps that directly define the cost value and its gradient with bilinear interpolation. }
    \vspace{-3ex}
    \label{fig_teb}
\end{figure}

We want to use these predictions in the Timed-Elastic-Band (TEB) planner \cite{rosmann2015planning, rosmann2017integrated}, but the original implementation only handles point obstacles. We chose to modify the TEB implementation to be able to handle grid representations (see Figure \ref{fig_teb}). TEB originally minimizes a linearly decreasing cost function: $\mathcal{C}_\mathrm{obst} = \mathrm{max}\left(0, 1 - d/d_0\right)$, where $d$ is the distance from the optimized pose to the closest obstacle and $d_0$ a predefined influence distance. The simple way to handle grid structures is to let TEB minimize the risk value at the current pose (interpolated from the closest grid values). Then we just need the grid values to represent a smooth cost function. In our case, we defined a linearly decreasing risk value similar to the original obstacle cost but with some modifications. First, we use a threshold $\tau_\mathrm{risk} = 0.4$ to extract risky area from the SOGM:

\begin{equation}
\label{eq3.1}
    \mathrm{R}^1_{k, i} = \mathrm{SOGM}_{k, i} > \tau_\mathrm{risk}  \: .
\end{equation}

\noindent Then we apply a 2D convolution to sum the polynomial decreasing cost from all pixels:

\begin{equation}
\label{eq3.2}
    \mathrm{R}^2_{k, i} =
      \sum\limits_{j}{\left(\mathcal{C}\left(i, j\right)^p \times \mathrm{R}^1_{k, j}\right)}  \: ,
\end{equation}

\noindent with $\mathcal{C}\left(i, j\right) = \mathrm{max}\left(0, 1 -  d\left(i, j\right) \times dl_\mathrm{2D} / d_0 \right)$, where $d\left(i, j\right)$ is the distance from pixel $i$ to pixel $j$ in the grid space, $p$ is explained later, and $d_0$ has the same meaning as in the original TEB, the influence distance of obstacles. This convolution diffuses the risk in space, but we also decided to diffuse the risk in time:

\begin{equation}
\label{eq3.3}
    \mathrm{R}^3_{k, i} =
      \sum\limits_{l}{\left(\mathcal{C}_\mathrm{t}\left(k, l\right)^p \times \mathrm{R}^2_{l, j}\right)}  \: ,
\end{equation}

\noindent with $\mathcal{C}_\mathrm{t}\left(k, l\right) = \mathrm{max}\left(0, 1 - \left\|t_k - t_l\right\| / \Delta_0 \right)$, where $t_{k/l}$ is the time at layer $k/l$, $p$ is explained later, and $\Delta_0 = 1$s is the influence range in time.

Summing risk this way means larger risk areas will have higher risk values. To even out the risk value for any risk area, we apply the same convolution but on normalized risk: 

\begin{equation}
\label{eq3.4}
    \mathrm{R}^4_{k, i} =
      \sum\limits_{l}{
      \sum\limits_{j}{
      \mathcal{C}_\mathrm{t}\left(k, l\right)^p 
      \left(\mathcal{C}\left(i, j\right)^p \left(\mathrm{R}^1_{l, j} / \mathrm{R}^3_{l, j}\right)
      \right)}
      }  \: .
\end{equation}

\noindent We put the linearly decreasing cost value to the power $p$, but we can retrieve a linearly decreasing diffusion by taking the power $1/p$ of this final cost:

\begin{equation}
\label{eq3.5}
    \mathrm{SRM}_{k, i} = \left(\mathrm{R}^4_{k, i}\right)^{\left(1/p\right)}  \: .
\end{equation}

\begin{figure}[t]
    \vspace{-2ex}
    \centering
    \includegraphics[width=0.999\columnwidth, keepaspectratio=true]{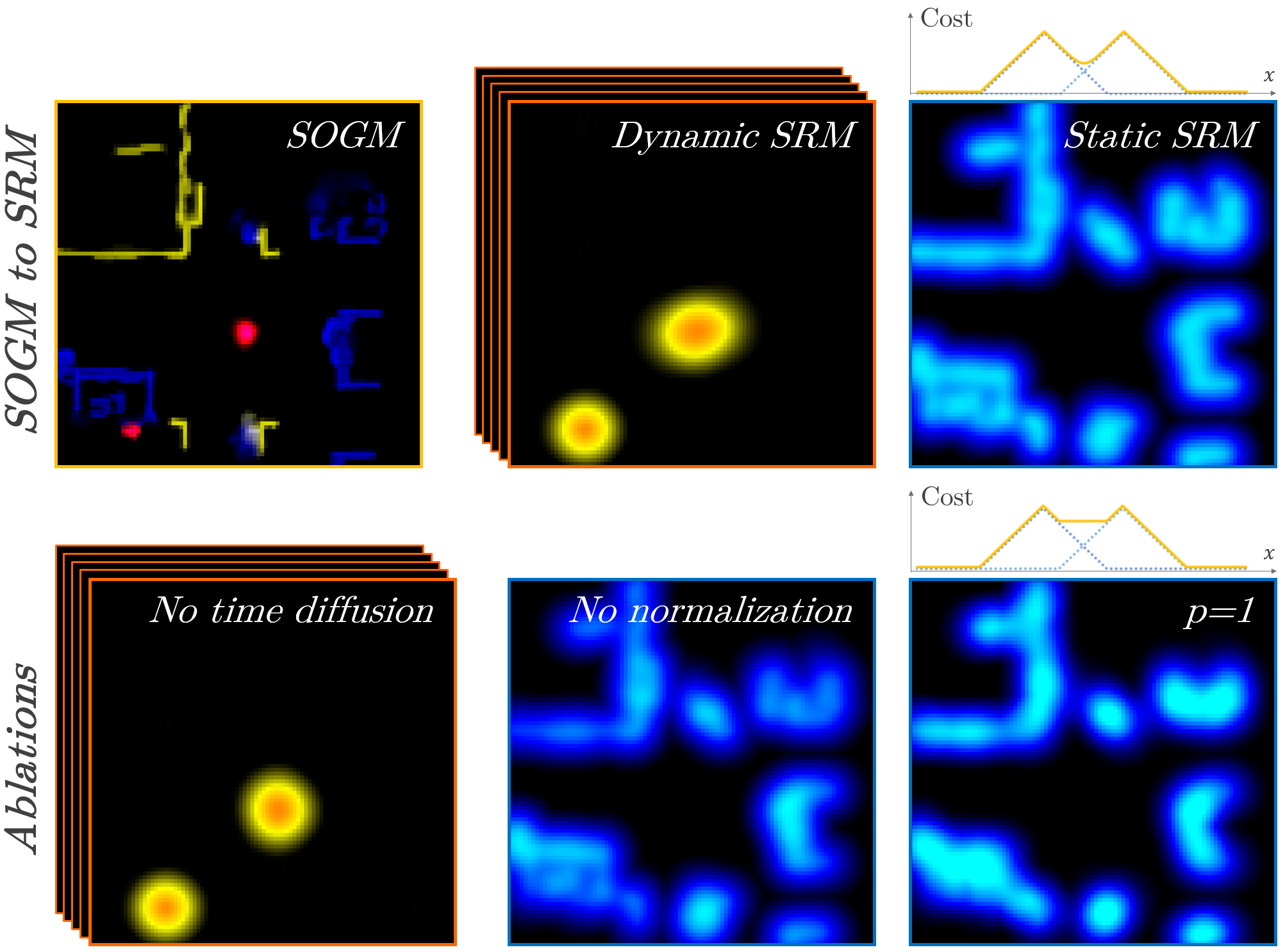}
    \caption{Conversion of SOGMs into decoupled static and dynamic SRMs. We show the impact of time diffusion, normalization, and parameter $p$. We show the effect of parameter $p$ in a small graph on top of the static SRM.}
    \label{fig_srm}
\end{figure}

\noindent This risk value behaves like a $p$-norm (see the small graphs in Figure \ref{fig_srm}), the higher $p$ is, the closer it is to the maximum value of the linear influence of each surrounding pixel. We use $p = 3$ in the following.

In addition to this new definition of SRM, we decouple the \textit{static} risk and \textit{dynamic} risk. The \textit{dynamic} risk is computed from the \textit{dynamic} channel of the SOGM, while the \textit{static} risk comes from the \textit{permanent} and \textit{movable} SOGM channels. Because the \textit{static} risk is the same at any time, it can be stored in a single 2D risk map. For convenience, we store it in the first layer of the SRM, while the rest of the SRM layers only contain the \textit{dynamic} risk. This decoupling allows us to have separate distance and weight parameters for both risks, and keep better control of the navigation behavior. The \textit{dynamic} risk is diffused in space and time defined as above with $d_0^{dyn}=1.2$m, and $\Delta_0 = 1$s and the static risk is diffused only in space with $d_0^{sta}=0.9$m.

If a pose time is too far in the future, it ignores the dynamic risk. TEB also allows the optimization of multiple trajectories for different homotopy classes. We keep this feature by creating estimated point obstacles at local maxima in the SOGM, ignored by the trajectory optimizer, but used for homotopy class computation.

\begin{figure*}[t]
    \centering
    \includegraphics[width=0.999\textwidth, keepaspectratio=true]{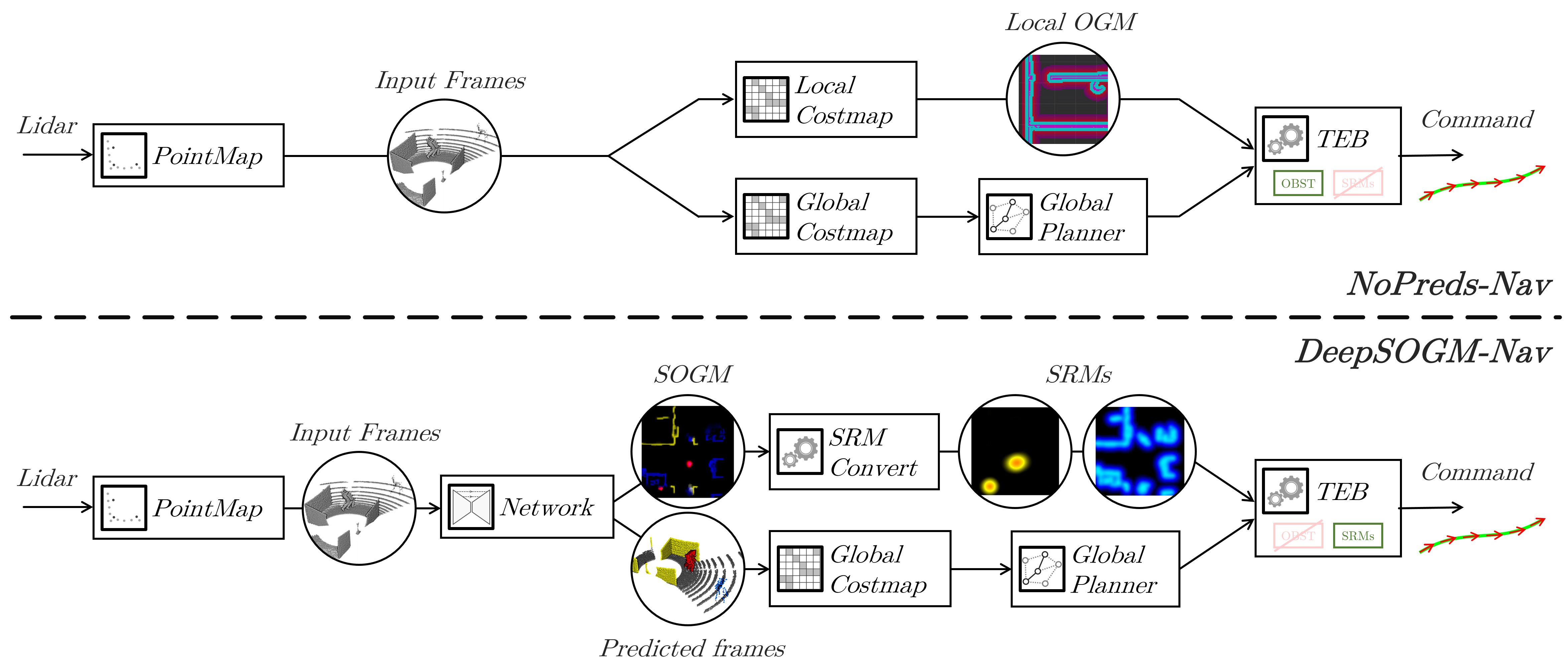}
    \caption{Illustration of the two navigation systems implemented on the simulated and the real robot. \textit{NoPreds-Nav} is a standard ROS navigation system using the regular TEB planner without predictions. \textit{DeepSOGM-Nav} integrates the network predictions.}
    \vspace{-3ex}
    \label{fig_nav}
\end{figure*}

In simulation, we have access to high computing power, with an Nvidia RTX 3090 GPU and an Intel i9-10980XE CPU @ 3.00GHz. In addition, we can slow the simulation time factor to reduce delay to virtually zero if we need. On the real robot, we are limited to a laptop configuration, with a much smaller GPU (Nvidia T1000) and a much slower CPU (i7-11800H @ 2.30GHz). However, with our current implementation and parameters (especially $dl_\mathrm{3D}=12$cm), we can run everything in real-time. The input frames arrive from PointMap with a delay of approximately $45$ms. The CPU pre-processing takes on average $47$ms and the GPU network computations $74$ms. Finally, $53$s are used for the SRM conversion. The total delay to get a new prediction varies between $230$ms ($10^\mathrm{th}$ percentile) and $320$ms ($90^\mathrm{th}$ percentile), with an average of $259$ms, which is totally acceptable. It means only the first few layers of the predictions are obsolete. To remain closer to the real configuration in our simulation experiments, we simulate this delay by waiting before publishing the network results to the rest of the pipeline.

%
%
%
%
%
%

\subsection{Standard Navigation System and Prediction Integration}

Our network predictions can easily be plugged into a standard navigation system. We use original ROS plugins for most of the navigation except for localization, which is performed with PointMap (adapted as a ROS plugin), and the local planner, which is our modified version of TEB. As shown in Figure \ref{fig_nav}, in the standard navigation pipeline, lidar frames are processed by PointMap to get the current robot pose. Then local and global costmaps are computed with the \textit{move\_base} ROS node. The global planner finds the optimal path to the goal and TEB follows this path while avoiding obstacles in the local costmap.

When using deep predictions, the subsampled and aligned frame from PointMap is sent to the network to be labeled, and to produce the SOGM, immediately converted into SRMs. The global costmap is computed with the labeled points and ignores dynamic obstacles. TEB tries to follow the global plan while avoiding high-risk areas in the SRMs. Note that we use the raw lidar frame for localization as opposed to \cite{thomas2021self}, where we used predicted frames, because our network needs three aligned input frames to be able to extract the current speed of dynamic points.

%
%
%
%
%
%
%
%
%
%
%
%

\section{Simulation Experiments}
\label{sec_simu}

In \cite{thomas2022learning}, we evaluated our Deep SOGM predictions on simulated data. We showed that our predictions could generalize to different types of actors, compared  them to  the predictions of other methods, and provided ablation studies. In this section, we complete these experiments with an evaluation of our navigation system. In particular, we compare the efficiency and safety of the TEB planner when using different types of SOGM predictions, or no prediction at all. We choose to conduct these experiments in simulation to allow large-scale testing, true metrics, and a repeatable controlled scenario for comparable results.

%
%
%
%
%
%

\subsection{Simulation Setup}

\begin{figure}[t]
    \centering
    \includegraphics[width=0.999\columnwidth, keepaspectratio=true]{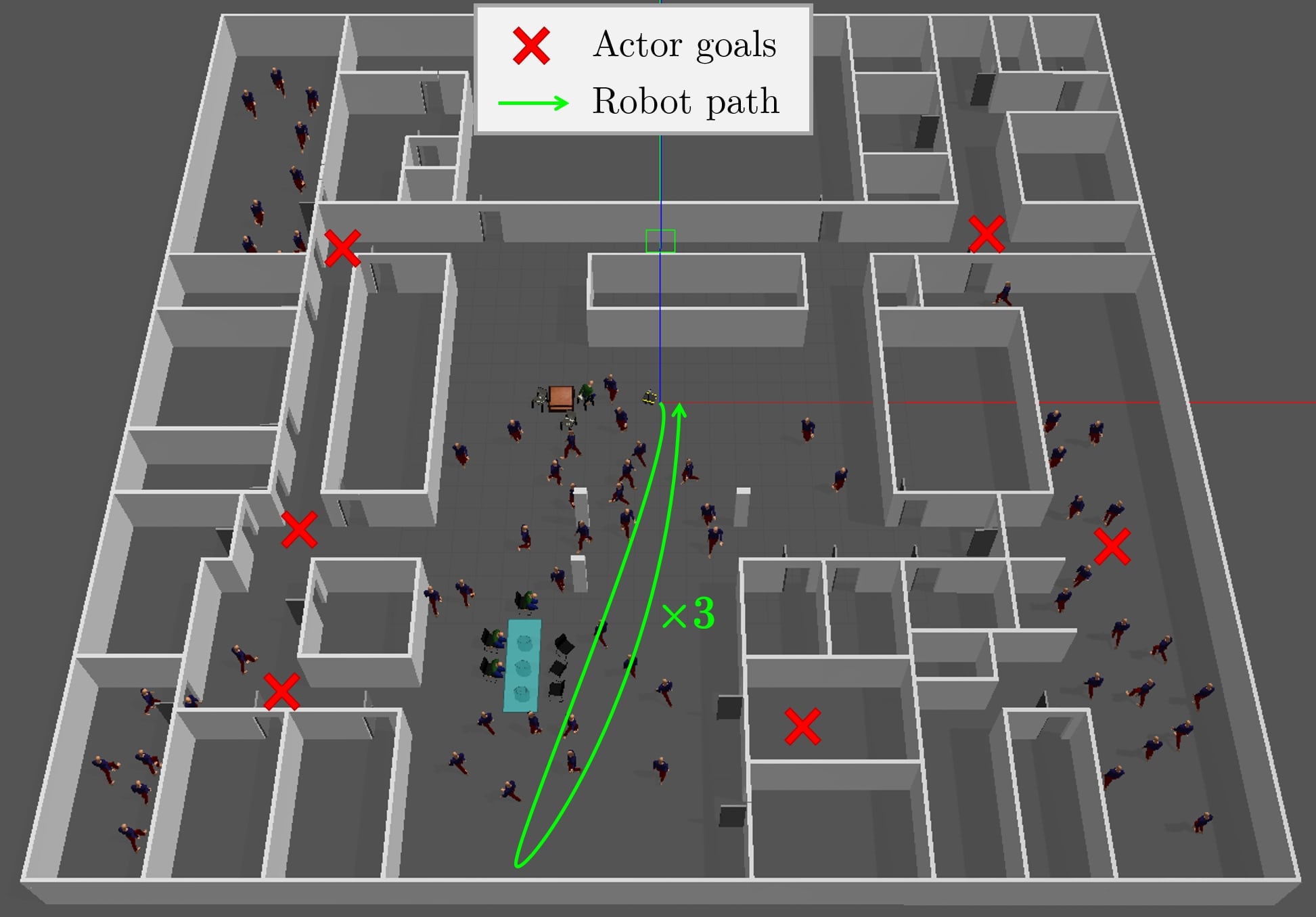}
    \caption{Simulation setup for our experiments. Actors are walking towards a goal randomly selected among the possible red cross locations. The robot navigates back and forth in the main atrium.}
    \vspace{-3ex}
    \label{fig_simusetup}
\end{figure}

We use the same Gazebo simulated environment as in \cite{thomas2021self, thomas2022learning} for our experiments. In this case, we designed a controlled experiment that could be repeated many times to get reliable results and fair comparisons between all methods. Tables and chairs are generated randomly in the space and a fixed number of Flow Followers \cite{thomas2022learning} are moving between a set of goals that we chose around the robot path, to force a lot of encounters. The robot is always asked to follow the same path, consisting of going across the main atrium a few times. See Figure \ref{fig_simusetup} for a visualization of this setup. 

In our experiments, we use metrics that are simple and intuitive. To measure the efficiency of the planner, we use the \textbf{Time to Reach the Final Goal} ($T_f$) $\downarrow$ in seconds. To measure the safety of the planner, we measure the distance from the center of the robot to the center of the closest dynamic actor (whose position is given by the simulator), and derive two metrics from it: the \textbf{Collision Ratio} (\%C) $\downarrow$, measuring the percentage of the total time during which the robot is in collision with an actor (distance smaller than $d_c = 0.4m$); and the \textbf{Risk Ratio} (\%R) $\downarrow$, measuring the proportion of the session during which the robot is in a risky area (distance smaller than $d_r = 1.0m$), which indicates when the robot is dangerously close to an actor. In addition to these main metrics, we provide four additional metrics providing more insight into the results:
\begin{itemize}
\item{average absolute speed (AAS) $\uparrow$ }
\item{percentage of time stopped (\%S) $\downarrow$ }
\item{average linear speed (ALS) $\uparrow$ }
\item{percentage of time going backwards (\%B) $\downarrow$ }
\end{itemize}

\noindent The AAS measures the robot's absolute speed in the horizontal $(x, y)$ plane, which is always positive, regardless of the direction, and is averaged across the session. On the contrary, ALS measures the speed with respect to the robot heading direction, which can be negative. The \%S and \%B metrics are measured with  absolute speed and linear speed respectively. The \%S is the proportion of time when the absolute speed is inferior to $0.1$ m.s$^{-1}$, while the \%B is the proportion of time when the linear speed is inferior to $-0.1$ m.s$^{-1}$.

%
%
%
%
%
%

\subsection{Planner Comparison Using Different Types of Predictions}

\begin{figure}[b]
    \centering
    \includegraphics[width=0.999\columnwidth, keepaspectratio=true]{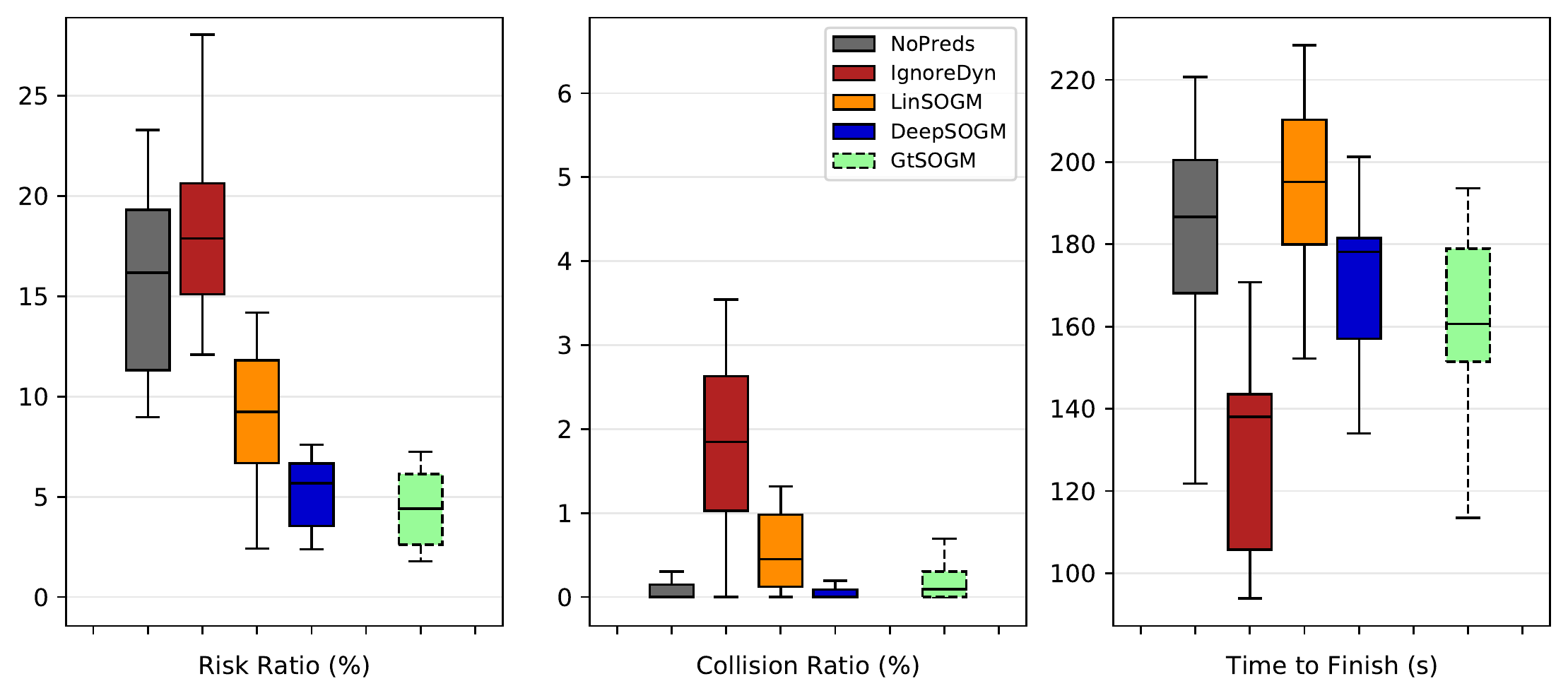}
    \caption{Evaluation of the Safety and Efficiency of the TEB planner with different types of predictions. For each type of prediction, results are collected over 20 different sessions.}
    \vspace{-3ex}
    \label{fig_comp}
\end{figure}

In our first experiment, we verify the benefit of our Deep-SOGM predictions for navigation. We thus keep the TEB planner and its parameters fixed and compare its performances when using:

\begin{itemize}
\item{\textit{NoPreds}: original version of the TEB ROS package, using local costmap pixels as obstacles.}
\item{\textit{IgnoreDyn}: idea from \cite{thomas2021self}, ignoring the dynamic obstacles, only using the static SOGM.}
\item{\textit{LinSOGM}: use actor current speeds (provided by the simulator), and extrapolate their positions linearly.}
\item{\textit{DeepSOGM}: our deep SOGM predictions. We use the network trained on Flow Followers from \cite{thomas2022learning}, and use it for inference here.}
\item{\textit{GtSOGM}: precompute actors’ movements in advance, to get the groundtruth future SOGM when navigating.}
\end{itemize}

\begin{figure}[t]
    \centering
    \includegraphics[width=0.75\columnwidth, keepaspectratio=true]{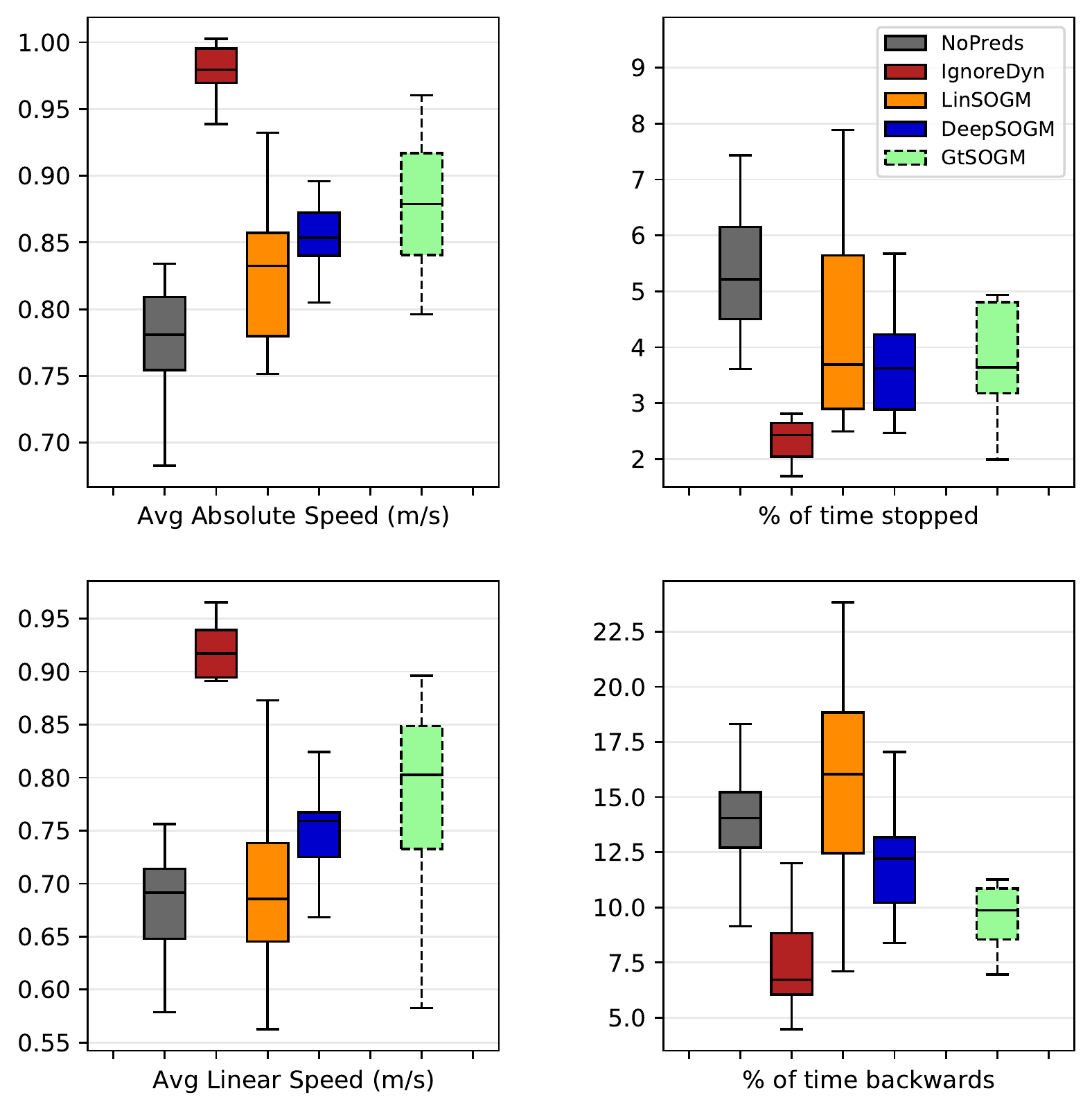}
    \caption{Additional metrics for the evaluation of the Safety and Efficiency of the TEB planner with different types of predictions.}
    \vspace{-3ex}
    \label{fig_comp_add}
\end{figure}

\noindent For a fair comparison, we enforce the same 250ms delay for the methods using SOGMs, to reflect what the real robot would be able to achieve. Note that for the \textit{GtSOGM} method, the actors will not react to the robot, as their movements are computed in advance. We find that this slight difference does not affect the results, because the FlowFollowers only try to avoid the robot if it is nearly colliding with them. For each method, we repeat the experiments 20 times to get a reliable average, std, and box plot. The results are compiled in Figure \ref{fig_comp} and Figure \ref{fig_comp_add} completes these results with additional metrics.

First, we look at TEB without predictions, \textit{NoPreds}, and see that it nearly never collides (\%C $< 0.5$). However, it often gets into risky situations and has to stop, therefore, it ends up being inefficient, with a longer time to finish. The first idea, \textit{IgnoreDyn}, proposed in \cite{thomas2021self}, is to ignore dynamic obstacles, hoping for the fact that they will avoid the robot on their own. Even if the Flow Followers are implemented to avoid the robot, they end up colliding often with it ($2$\%C on average). We argue that people would be better at avoiding the robot, but it would be risky to rely solely on people's reactions, and it would probably lead to many collisions. We notice that this planner is the fastest ($T_f$ < $140$s on average), and rarely stops, because it goes straight to the goal, without avoiding the dynamic obstacles. Then we evaluate a planner using the linear extrapolation of the actor's current speeds, \textit{LinSOGM}. This method gives the robot a sense of the future movement of the actors, which leads to a reduced time in risky and collision areas. But it is not ideal yet, because the robot anticipates on an approximate linear prediction, and has to readjust many times. It particularly affects the efficiency (time to finish) and is even worse than the regular TEB. Our version of TEB, \textit{DeepSOGM}, performs well compared to the other methods, with close to zero collision (\%C $< 0.5$), the shortest time in risk areas ($5$\%R on average), and a relatively fast finishing time. Eventually, we compare the performances to the SOGMs using the actual groundtruth future provided by the simulator, and this is probably the most impressive result. Our performances are extremely close to the performances of a robot that could actually predict the future. We believe this result is more useful than a comparison to other OGM prediction methods, which would be complex to implement in our pipeline without modifying several other components. It lets us believe that our prediction method, in itself, is strong enough to provide SOGMs of sufficient quality for navigation and that further improvements would probably be achieved by upgrading other components (SRM conversion, planner, etc.), or finding ways to prolong the prediction time horizon.

In our \textbf{supplementary video}, we show the robot navigating with different types of prediction, first in rviz view, where predictions can be visualized, and then in a schematic bird-eye view where the difference in trajectories between \textit{NoPreds} and \textit{DeepSOGM} can be seen.

%
%
%
%
%
%

\subsection{Ablation Studies of the Planner Using Deep-SOGM}

\begin{figure}[t]
    \centering
    \includegraphics[width=0.999\columnwidth, keepaspectratio=true]{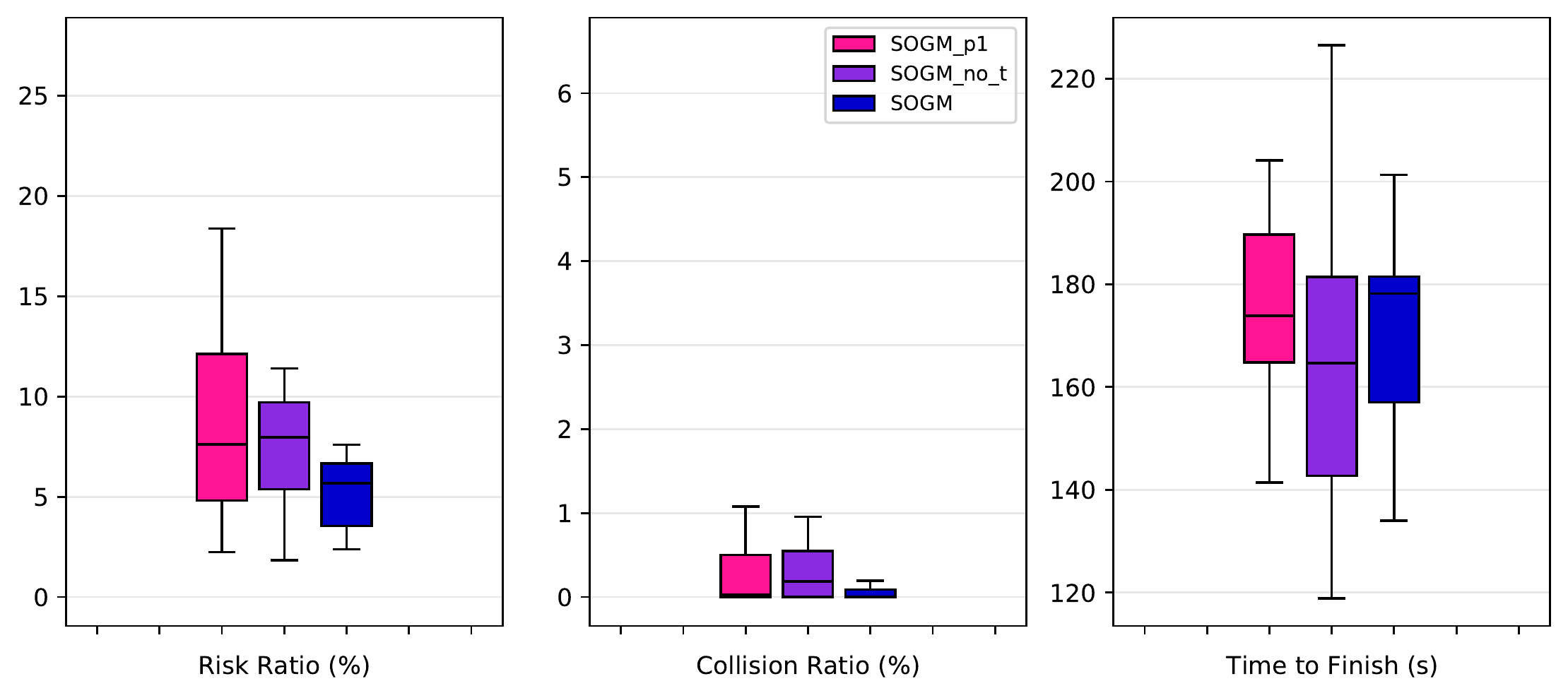}
    \caption{Evaluation of the Safety and Efficiency of our DeepSOGM TEB planner without some key components for SOGM-SRM conversion. For each ablation study, results are collected over 20 different sessions.}
    \vspace{-3ex}
    \label{fig_ablation}
\end{figure}

In this second experiment, we use the same protocol to compare three versions of our Planner using our Deep-SOGM predictions. We show how some of the key choices made for the SOGM-to-SRM conversion affect the performances in terms of safety and efficiency in Figure \ref{fig_ablation}. First, we measure the performances when computing SRM with $p=1$. The navigation is riskier and less efficient because the risk function is not well defined in the space between obstacles. Then we remove the diffusion of the risk in the time dimension, one of the additions made in this work. In that case, the robot can plan trajectories closer to the back or the front of moving actors, which naturally increases the time spent in risky areas. It allows the robot to go faster in some cases, but also means it will end up more often in collision situations, where it has to stop and reverse. Therefore, the distribution of time to finish is more spread out for this method. In both cases, our final version of the planner has better performance.

%
%
%
%
%
%
%
%
%
%
%
%

\section{Real-world Experiments}
\label{sec_real}

A main goal of this paper is to validate that our algorithms generalize to real-world indoor navigation. In this section, we analyze our network predictions and our navigation system using real data. First, we can study the network predictions on their own, in a similar fashion as \cite{thomas2022learning}, by comparing the network predictions to the data annotated by our automated pipeline. We evaluate how the predictions can improve over time, in a lifelong learning manner, and how adding simulated data to the training set impacts the results. Then we analyze our navigation system. In the real world, it is hard to reproduce multiple navigation experiments as we could do in simulation, but we show, with anecdotal examples, that the conclusions from simulated experiments generalize to real data. Eventually, we compile the data collected during our experiments to provide a new 3D lidar dataset with indoor pedestrians for the robotics community.

%
%
%
%
%
%

\subsection{Real-world Setup}

\begin{figure}[t]
    \centering
    \includegraphics[width=0.999\columnwidth, keepaspectratio=true]{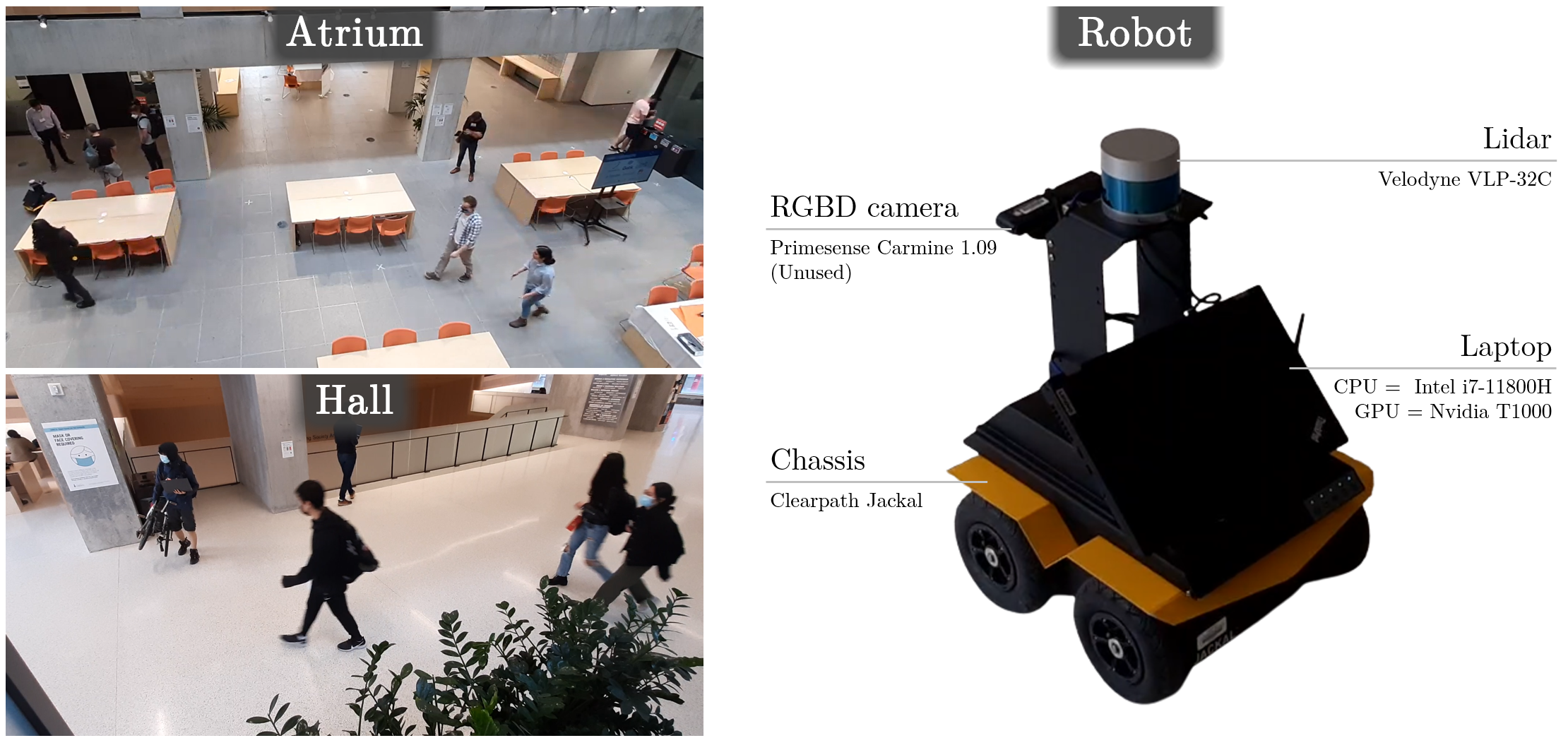}
    \caption{Our real robot setup and experiment spaces. In this work, we only use the lidar sensor and perform most of the computations on a laptop fixed to the robot.}
    \vspace{-3ex}
    \label{fig_realsetup}
\end{figure}

For the real-world experiments, we use a Clearpath Jackal robot, shown in Figure \ref{fig_realsetup}. It is a small field robotics research platform, with an onboard computer, GPS, and IMU fully integrated with ROS. In this work, we use a single 3D sensor: a Velodyne VLP-32C lidar sensor. An RGBD camera is mounted on the robot, but we do not use it for our experiments. To this platform, we add a laptop computer with an Intel CPU (i7-11800H @ 2.30GHz) and an Nvidia GPU (Nvidia T1000 4GB). Most of the computations (localization, planning, inference) are performed on the laptop. Only basic tasks (Velodyne processing, low-level control) are performed on the onboard Jackal computer.

In the real world, it is hard to reproduce multiple experiments as we could do in simulation. On the one hand, if you choose to navigate ``in-the-wild'' in a space where people are not told to behave in a particular way around the robot, the navigation conditions from one session to another will be totally different depending on how people react to the robot or try to confuse it. On the other hand, if you choose to have a controlled experiment, where people are told to act in a particular way, you can assume that the navigation conditions will be roughly similar, but you limit yourself to over-simplified situations and behaviors, and cannot be sure that the results will generalize well to any circumstances.

Having verified our navigation system performances in simulation, another thorough study on our real-world platform is not crucial, and we decided to conduct both controlled and in-the-wild experiments to validate that our robots behave as intended and to confirm the results from Section \ref{sec_simu}. This is why we perform experiments in two different spaces of the same building. The \textbf{Atrium} has several tables and chairs that are often moved and configured differently depending on the occasion. In this space, students usually come to work and thus dynamic obstacles are not very common, except during specific events. This space was used to conduct more controlled experiments. The main \textbf{Hall} of the building has a big entrance, stairs, and elevator that lead to classrooms, and large open spaces without tables, where students come and go depending on where they are heading. During peak hours, this space can be crowded with dynamic obstacles, which was ideal for our in-the-wild experiments. Pictures of both spaces are shown in Figure \ref{fig_realsetup}.

%
%
%
%
%
%

\subsection{Real Data Collection and Automated Annotation}

\begin{figure}[b]
    \centering
    \includegraphics[width=0.999\columnwidth, keepaspectratio=true]{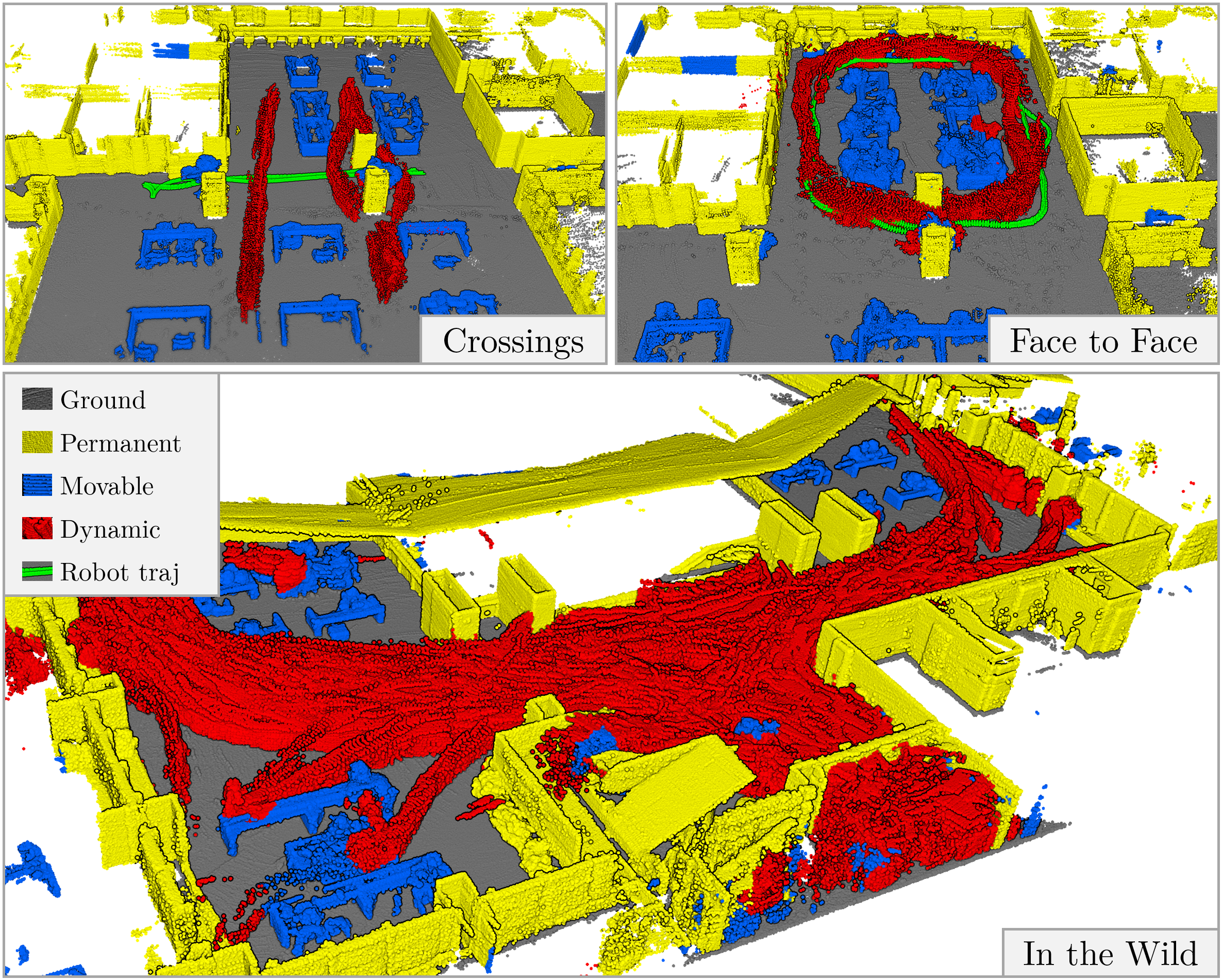}
    \caption{Different parts of our 3D lidar point cloud dataset, annotated by our automated pipeline. We see controlled scenarios with the robot trajectory in green, and the \textit{dynamic} points in red. A crowded in-the-wild session is also given as an example.}
    \vspace{-3ex}
    \label{fig_realdata}
\end{figure}

Our first real experiments were conducted in the Atrium because it is ideal for controlled scenarios. With low traffic, and people mostly sitting for long periods, it is less likely that unexpected circumstances arise there. Following our automated annotation pipeline, we first conducted a mapping session, obtaining the initial map shown in Figure \ref{fig_annot}. Mapping and refinement were done by driving the robot manually for convenience, but could also have been done with the \textit{NoPreds-Nav}, with PointMap in SLAM mode. We started by collecting a first batch of 9 sessions in the space without interfering. Therefore, only a handful of dynamic obstacles were encountered. A second batch of 10 sessions was collected in a controlled manner, where a person was asked to cross the robot's path perpendicularly at three different predefined points (shown in Figure \ref{fig_realdata}). The third batch of 6 sessions was collected with a focus on face-to-face encounters. The robot was asked to navigate along a looping trajectory and a person was told to walk along the same loop in the opposite direction (See Figure \ref{fig_realdata}). Finally, we collected 15 additional sessions during a conference that was organized in the building. For these sessions, the layout of the Atrium was different and more people were present in the space without any instructions on how to behave around the robot.

For more ``in-the-wild'' results, we also collected data in the main hall of the building. After the initial mapping and the refinement runs, we collected 38 sessions, at different hours on different days, alternating between crowded times (See Figure \ref{fig_realdata}) and more calm moments. The sessions collected in this space are not organized in a particular order, because they all were collected without any instructions given to the people in the space. In these sessions (and also in the last batch of sessions in the atrium), we noticed several people trying to mess with the robot by acting in unexpected ways and sometimes had to stop the robot ourselves to avoid collisions. Ideally, we would like the robot to be able to predict any kind of behavior, even the disrupting ones, which is why we keep these sessions in the dataset. Most of the sessions were collected using the \textit{NoPreds-Nav}, and some later sessions were collected with the \textit{DeepSOGM-Nav}, using a network trained on earlier sessions. From a data collection and annotation point of view, this does not really matter. A different robot's behavior may induce different reactions from the people around it, but these cases rarely happen. The collected sessions are listed with details in Tables \ref{Table_UTIn3D-A} and \ref{Table_UTIn3D-H}. Our open dataset is called \textbf{UofT-Indoor-3D (UTIn3D)} and is available for the community to use. We share the lidar frames, the trajectories computed by PointMap, and our annotations. 3D lidar datasets with crowds of indoor pedestrians are not very common, and we hope that UTIn3D will be beneficial for the community.

\begin{table}[t]
\caption{Description of all the sessions in UTIn3D-Atrium Dataset. For each session we specify the time, the number of frames ($N_f$), the number of points (in millions), the percentage of frames containing dynamic points ($\mathrm{DynF}$), and the percentage of dynamic points ($\mathrm{DynP}$). We highlight crowded sessions in bold when $\mathrm{DynF}>50\%$ or $\mathrm{DynP}>5\%$.}
\setlength\tabcolsep{0.5pt}
\begin{footnotesize}
\begin{center}
\begin{tabular}{ C{0.4cm} C{2.5cm} C{0.5cm} C{0.5cm} C{0.9cm} C{0.9cm} C{0.9cm} C{0.9cm} C{0.9cm} }

 & \multicolumn{8}{c}{\textbf{UTIn3D-Atrium}} \\
\Xhline{2\arrayrulewidth}
 & Date & \multicolumn{2}{c}{Tr/Va} & Time & $N_f$ & Mpts  & $\mathrm{DynF}$ & $\mathrm{DynP}$  \TBstrut\\
\Xhline{2\arrayrulewidth}

\multirow{9}{*}{\rot{\textbf{UTIn3D-A1}}} &	2021-12-10\_12-53-37 & 	\ding{51} & 	 & 	0:04:21 & 	$2610$ & 	$271.0$ & 	$43.3\%$ & 	$1.9\%$ 	\Tstrut\\
 & 	2021-12-10\_13-06-09 & 	 & 	\ding{51} & 	0:03:37 & 	$2166$ & 	$224.8$ & 	$6.8\%$ & 	$.5\%$ 	\\
 & 	2021-12-10\_13-17-29 & 	 & 	\ding{51} & 	0:04:22 & 	$2617$ & 	$277.1$ & 	$.0\%$ & 	$.1\%$ 	\\
 & 	2021-12-10\_13-26-07 & 	\ding{51} & 	 & 	0:02:58 & 	$1782$ & 	$189.3$ & 	$21.3\%$ & 	$1.2\%$ 	\\
 & 	2021-12-10\_13-32-10 & 	\ding{51} & 	 & 	0:04:12 & 	$2519$ & 	$267.9$ & 	$39.2\%$ & 	$3.8\%$ 	\\
 & 	2021-12-13\_18-16-27 & 	\ding{51} & 	 & 	0:02:54 & 	$1743$ & 	$183.0$ & 	$29.4\%$ & 	$1.2\%$ 	\\
 & 	2021-12-13\_18-22-11 & 	\ding{51} & 	 & 	0:04:17 & 	$2568$ & 	$275.8$ & 	$19.9\%$ & 	$1.0\%$ 	\\
 & 	2021-12-15\_19-09-57 & 	\ding{51} & 	 & 	0:02:31 & 	$1511$ & 	$159.9$ & 	$17.6\%$ & 	$1.1\%$ 	\\
 & 	2021-12-15\_19-13-03 & 	\ding{51} & 	 & 	0:03:43 & 	$2229$ & 	$238.3$ & 	$22.5\%$ & 	$1.0\%$ 	\Bstrut\\
 \hline
\multirow{10}{*}{\rot{\textbf{UTIn3D-A2}}} &	2022-01-18\_10-38-28 & 	\ding{51} & 	 & 	0:03:35 & 	$2151$ & 	$228.2$ & 	$49.9\%$ & 	$3.5\%$ 	\Tstrut\\
 & 	2022-01-18\_10-42-54 & 	\ding{51} & 	 & 	0:03:38 & 	$2180$ & 	$231.1$ & 	$30.7\%$ & 	$2.0\%$ 	\\
 & 	2022-01-18\_10-47-07 & 	\ding{51} & 	 & 	0:01:06 & 	$664$ & 	$68.5$ & 	$22.1\%$ & 	$1.1\%$ 	\\
 & 	2022-01-18\_10-48-42 & 	\ding{51} & 	 & 	0:03:51 & 	$2306$ & 	$244.6$ & 	$21.2\%$ & 	$1.2\%$ 	\\
 & 	2022-01-18\_10-53-28 & 	\ding{51} & 	 & 	0:03:43 & 	$2232$ & 	$236.8$ & 	$50.6\%$ & 	$4.0\%$ 	\\
 & 	2022-01-18\_10-58-05 & 	\ding{51} & 	 & 	0:03:42 & 	$2216$ & 	$235.1$ & 	$37.9\%$ & 	$3.6\%$ 	\\
 & 	2022-01-18\_11-02-28 & 	\ding{51} & 	 & 	0:03:36 & 	$2161$ & 	$229.4$ & 	$25.4\%$ & 	$1.4\%$ 	\\
 & 	2022-01-18\_11-11-03 & 	\ding{51} & 	 & 	0:03:53 & 	$2333$ & 	$247.4$ & 	$47.8\%$ & 	$3.6\%$ 	\\
 & 	2022-01-18\_11-15-40 & 	\ding{51} & 	 & 	0:03:32 & 	$2126$ & 	$225.5$ & 	$31.0\%$ & 	$1.8\%$ 	\\
 & 	2022-01-18\_11-20-21 & 	\ding{51} & 	 & 	0:04:02 & 	$2422$ & 	$257.5$ & 	$35.3\%$ & 	$1.8\%$ 	\Bstrut\\
 \hline
\multirow{6}{*}{\rot{\textbf{UTIn3D-A3}}} &	2022-02-25\_18-19-12 & 	\ding{51} & 	 & 	0:03:11 & 	$1909$ & 	$201.2$ & 	$27.3\%$ & 	$1.4\%$ 	\Tstrut\\
 & 	2022-02-25\_18-24-30 & 	\ding{51} & 	 & 	0:03:07 & 	$1869$ & 	$197.9$ & 	$28.0\%$ & 	$1.2\%$ 	\\
 & 	2022-02-25\_18-29-18 & 	\ding{51} & 	 & 	0:03:24 & 	$2041$ & 	$215.9$ & 	$23.1\%$ & 	$1.0\%$ 	\\
 & 	2022-03-01\_22-01-13 & 	\ding{51} & 	 & 	0:03:08 & 	$1879$ & 	$199.5$ & 	$57.2\%$ & 	$2.6\%$ 	\\
 & 	2022-03-01\_22-06-28 & 	\ding{51} & 	 & 	0:03:42 & 	$2222$ & 	$236.2$ & 	$27.7\%$ & 	$1.3\%$ 	\\
 & 	2022-03-01\_22-25-19 & 	\ding{51} & 	 & 	0:03:44 & 	$2243$ & 	$238.8$ & 	$43.2\%$ & 	$2.8\%$ 	\Bstrut\\
 \hline
\multirow{15}{*}{\rot{\textbf{UTIn3D-A4}}} &	2022-05-20\_12-47-48 & 	 & 	\ding{51} & 	0:03:02 & 	$1820$ & 	$190.3$ & 	$\mathbf{69.0\%}$ & 	$\mathbf{7.7\%}$ 	\Tstrut\\
 & 	2022-05-20\_12-54-23 & 	\ding{51} & 	 & 	0:03:01 & 	$1815$ & 	$191.0$ & 	$\mathbf{67.0\%}$ & 	$\mathbf{5.2\%}$ 	\\
 & 	2022-05-20\_12-58-26 & 	\ding{51} & 	 & 	0:03:37 & 	$2171$ & 	$226.3$ & 	$\mathbf{59.8\%}$ & 	$\mathbf{8.3\%}$ 	\\
 & 	2022-05-31\_14-45-53 & 	\ding{51} & 	 & 	0:02:16 & 	$1363$ & 	$143.5$ & 	$44.5\%$ & 	$3.5\%$ 	\\
 & 	2022-05-31\_16-25-23 & 	\ding{51} & 	 & 	0:02:53 & 	$1736$ & 	$184.3$ & 	$47.1\%$ & 	$3.0\%$ 	\\
 & 	2022-05-31\_16-29-56 & 	\ding{51} & 	 & 	0:02:52 & 	$1717$ & 	$182.4$ & 	$60.6\%$ & 	$4.5\%$ 	\\
 & 	2022-05-31\_16-35-32 & 	\ding{51} & 	 & 	0:01:49 & 	$1094$ & 	$115.5$ & 	$\mathbf{67.8\%}$ & 	$\mathbf{5.8\%}$ 	\\
 & 	2022-05-31\_16-38-34 & 	\ding{51} & 	 & 	0:02:00 & 	$1196$ & 	$126.6$ & 	$31.5\%$ & 	$2.3\%$ 	\\
 & 	2022-05-31\_18-33-02 & 	\ding{51} & 	 & 	0:02:01 & 	$1215$ & 	$129.0$ & 	$11.0\%$ & 	$.6\%$ 	\\
 & 	2022-05-31\_19-34-18 & 	 & 	\ding{51} & 	0:01:48 & 	$1082$ & 	$114.9$ & 	$33.3\%$ & 	$1.2\%$ 	\\
 & 	2022-05-31\_19-37-08 & 	\ding{51} & 	 & 	0:02:27 & 	$1467$ & 	$155.4$ & 	$77.8\%$ & 	$4.9\%$ 	\\
 & 	2022-05-31\_19-40-52 & 	 & 	\ding{51} & 	0:02:50 & 	$1702$ & 	$180.6$ & 	$54.1\%$ & 	$2.8\%$ 	\\
 & 	2022-05-31\_19-44-52 & 	\ding{51} & 	 & 	0:01:58 & 	$1177$ & 	$124.8$ & 	$51.3\%$ & 	$3.0\%$ 	\\
 & 	2022-05-31\_19-47-52 & 	\ding{51} & 	 & 	0:01:59 & 	$1194$ & 	$127.0$ & 	$50.6\%$ & 	$2.6\%$ 	\\
 & 	2022-05-31\_19-51-14 & 	\ding{51} & 	 & 	0:01:58 & 	$1184$ & 	$125.5$ & 	$24.9\%$ & 	$1.3\%$ 	\Bstrut\\
\Xhline{2\arrayrulewidth}
 & 	Total & 	35 & 	5 & 	2:04:19 & 	$74632$ & 	$7897.9$ & 	$37.7\%$ & 	$2.6\%$ 	\TBstrut\\
\Xhline{2\arrayrulewidth}

\end{tabular}
\end{center}
\end{footnotesize}
\label{Table_UTIn3D-A} 
\vspace{-3ex}
\end{table}

We do not have quantitative measurements of the quality of the annotation on real data, but we can observe the results qualitatively. For the most part, the annotation quality is very good. The different classes are quite well split, with only a few leaks from one class to another, for example where people are close to tables and then moving away, as we can see in Figure \ref{fig_realdata}. This type of mistake does not affect our navigation system so much as most encounters are happening far away from static objects like tables. Examples of annotated SOGMs can be seen in Figure \ref{fig_sogmcomp}.

\begin{table}[t]
\caption{Description of all the sessions in UTIn3D-Hall Dataset. For each session we specify the time, the number of frames ($N_f$), the number of points (in millions), the percentage of frames containing dynamic points ($\mathrm{DynF}$), and the percentage of dynamic points ($\mathrm{DynP}$). We highlight crowded sessions in bold when $\mathrm{DynF}>50\%$ or $\mathrm{DynP}>5\%$.}
\setlength\tabcolsep{0.5pt}
\begin{footnotesize}
\begin{center}
\begin{tabular}{ C{0.4cm} C{2.5cm} C{0.5cm} C{0.5cm} C{0.9cm} C{0.9cm} C{0.9cm} C{0.9cm} C{0.9cm} }

 & \multicolumn{8}{c}{\textbf{UTIn3D-Hall}}  \\
\Xhline{2\arrayrulewidth}
 & Date & \multicolumn{2}{c}{Tr/Va} & Time & $N_f$ & Mpts  & $\mathrm{DynF}$ & $\mathrm{DynP}$  \TBstrut\\
\Xhline{2\arrayrulewidth}

\multirow{38}{*}{\rot{\textbf{UTIn3D-H}}} &	2022-03-08\_21-02-28 & 	\ding{51} & 	 & 	0:04:04 & 	$2445$ & 	$257.2$ & 	$\mathbf{93\%}$ & 	$\mathbf{13.3\%}$ 	\Tstrut\\
 & 	2022-03-08\_21-08-04 & 	\ding{51} & 	 & 	0:02:24 & 	$1446$ & 	$152.5$ & 	$\mathbf{59\%}$ & 	$\mathbf{5.1\%}$ 	\\
 & 	2022-03-08\_22-19-08 & 	\ding{51} & 	 & 	0:03:27 & 	$2070$ & 	$219.9$ & 	$54\%$ & 	$4.1\%$ 	\\
 & 	2022-03-08\_22-24-22 & 	\ding{51} & 	 & 	0:02:28 & 	$1481$ & 	$156.3$ & 	$44\%$ & 	$3.8\%$ 	\\
 & 	2022-03-09\_15-55-10 & 	\ding{51} & 	 & 	0:02:40 & 	$1600$ & 	$168.9$ & 	$50\%$ & 	$3.8\%$ 	\\
 & 	2022-03-09\_15-58-56 & 	 & 	\ding{51} & 	0:03:24 & 	$2042$ & 	$217.1$ & 	$30\%$ & 	$1.7\%$ 	\\
 & 	2022-03-09\_16-03-21 & 	 & 	\ding{51} & 	0:02:51 & 	$1715$ & 	$180.6$ & 	$\mathbf{72\%}$ & 	$\mathbf{12.6\%}$ 	\\
 & 	2022-03-09\_16-07-11 & 	\ding{51} & 	 & 	0:02:29 & 	$1492$ & 	$157.1$ & 	$\mathbf{92\%}$ & 	$\mathbf{7.3\%}$ 	\\
 & 	2022-03-16\_16-05-29 & 	\ding{51} & 	 & 	0:02:57 & 	$1773$ & 	$186.4$ & 	$\mathbf{92\%}$ & 	$\mathbf{20.0\%}$ 	\\
 & 	2022-03-16\_20-05-22 & 	\ding{51} & 	 & 	0:02:57 & 	$1772$ & 	$185.9$ & 	$\mathbf{93\%}$ & 	$\mathbf{14.8\%}$ 	\\
 & 	2022-03-16\_20-13-08 & 	\ding{51} & 	 & 	0:03:17 & 	$1968$ & 	$206.5$ & 	$\mathbf{86\%}$ & 	$\mathbf{8.8\%}$ 	\\
 & 	2022-03-16\_21-21-35 & 	\ding{51} & 	 & 	0:03:30 & 	$2097$ & 	$222.9$ & 	$46\%$ & 	$2.7\%$ 	\\
 & 	2022-03-16\_21-28-09 & 	\ding{51} & 	 & 	0:02:02 & 	$1226$ & 	$129.4$ & 	$33\%$ & 	$2.8\%$ 	\\
 & 	2022-03-22\_14-04-53 & 	\ding{51} & 	 & 	0:01:15 & 	$745$ & 	$78.4$ & 	$\mathbf{81\%}$ & 	$\mathbf{10.2\%}$ 	\\
 & 	2022-03-22\_14-07-26 & 	\ding{51} & 	 & 	0:02:37 & 	$1572$ & 	$166.5$ & 	$\mathbf{93\%}$ & 	$\mathbf{8.9\%}$ 	\\
 & 	2022-03-22\_14-12-20 & 	 & 	\ding{51} & 	0:02:42 & 	$1625$ & 	$172.1$ & 	$\mathbf{54\%}$ & 	$\mathbf{5.6\%}$ 	\\
 & 	2022-03-22\_15-05-20 & 	\ding{51} & 	 & 	0:02:54 & 	$1737$ & 	$183.7$ & 	$\mathbf{92\%}$ & 	$\mathbf{12.2\%}$ 	\\
 & 	2022-03-22\_15-09-02 & 	\ding{51} & 	 & 	0:02:34 & 	$1539$ & 	$163.1$ & 	$\mathbf{57\%}$ & 	$\mathbf{6.0\%}$ 	\\
 & 	2022-03-22\_15-12-23 & 	\ding{51} & 	 & 	0:02:29 & 	$1492$ & 	$158.2$ & 	$57\%$ & 	$4.9\%$ 	\\
 & 	2022-03-22\_16-04-06 & 	\ding{51} & 	 & 	0:02:59 & 	$1789$ & 	$188.4$ & 	$\mathbf{100\%}$ & 	$\mathbf{18.7\%}$ 	\\
 & 	2022-03-22\_16-08-09 & 	 & 	\ding{51} & 	0:02:42 & 	$1622$ & 	$171.3$ & 	$\mathbf{95\%}$ & 	$\mathbf{11.3\%}$ 	\\
 & 	2022-03-28\_14-53-33 & 	 & 	\ding{51} & 	0:02:30 & 	$1502$ & 	$159.4$ & 	$25\%$ & 	$1.5\%$ 	\\
 & 	2022-03-28\_14-57-17 & 	\ding{51} & 	 & 	0:02:43 & 	$1635$ & 	$173.2$ & 	$53\%$ & 	$4.1\%$ 	\\
 & 	2022-03-28\_15-00-42 & 	\ding{51} & 	 & 	0:02:40 & 	$1603$ & 	$169.8$ & 	$\mathbf{66\%}$ & 	$\mathbf{5.4\%}$ 	\\
 & 	2022-03-28\_15-04-24 & 	\ding{51} & 	 & 	0:02:32 & 	$1520$ & 	$161.0$ & 	$42\%$ & 	$3.9\%$ 	\\
 & 	2022-03-28\_16-56-52 & 	 & 	\ding{51} & 	0:00:55 & 	$554$ & 	$58.5$ & 	$58\%$ & 	$3.7\%$ 	\\
 & 	2022-03-28\_17-03-29 & 	\ding{51} & 	 & 	0:01:36 & 	$959$ & 	$100.5$ & 	$\mathbf{100\%}$ & 	$\mathbf{35.7\%}$ 	\\
 & 	2022-03-28\_17-07-19 & 	\ding{51} & 	 & 	0:01:42 & 	$1025$ & 	$108.5$ & 	$\mathbf{83\%}$ & 	$\mathbf{9.0\%}$ 	\\
 & 	2022-03-28\_17-10-13 & 	\ding{51} & 	 & 	0:02:36 & 	$1564$ & 	$165.5$ & 	$\mathbf{53\%}$ & 	$\mathbf{5.5\%}$ 	\\
 & 	2022-03-28\_21-57-36 & 	\ding{51} & 	 & 	0:02:35 & 	$1550$ & 	$164.1$ & 	$\mathbf{71\%}$ & 	$\mathbf{5.5\%}$ 	\\
 & 	2022-03-28\_22-02-15 & 	\ding{51} & 	 & 	0:02:50 & 	$1700$ & 	$179.8$ & 	$\mathbf{78\%}$ & 	$\mathbf{7.8\%}$ 	\\
 & 	2022-04-01\_14-00-06 & 	 & 	\ding{51} & 	0:02:19 & 	$1387$ & 	$146.6$ & 	$59\%$ & 	$5.0\%$ 	\\
 & 	2022-04-01\_14-03-50 & 	\ding{51} & 	 & 	0:02:48 & 	$1681$ & 	$177.3$ & 	$\mathbf{94\%}$ & 	$\mathbf{13.5\%}$ 	\\
 & 	2022-04-01\_14-53-42 & 	\ding{51} & 	 & 	0:01:29 & 	$891$ & 	$94.5$ & 	$75\%$ & 	$3.5\%$ 	\\
 & 	2022-04-01\_14-57-35 & 	 & 	\ding{51} & 	0:02:17 & 	$1376$ & 	$145.8$ & 	$44\%$ & 	$2.3\%$ 	\\
 & 	2022-04-01\_15-01-18 & 	\ding{51} & 	 & 	0:04:33 & 	$2734$ & 	$288.3$ & 	$\mathbf{99\%}$ & 	$\mathbf{29.1\%}$ 	\\
 & 	2022-04-01\_15-06-55 & 	 & 	\ding{51} & 	0:03:21 & 	$2008$ & 	$211.3$ & 	$\mathbf{92\%}$ & 	$\mathbf{15.1\%}$ 	\\
 & 	2022-04-01\_15-11-29 & 	 & 	\ding{51} & 	0:02:36 & 	$1566$ & 	$165.5$ & 	$\mathbf{54\%}$ & 	$\mathbf{5.1\%}$ 	\Bstrut\\
\Xhline{2\arrayrulewidth}
 & 	Total & 	28 & 	10 & 	1:40:47 & 	$60503$ & 	$6391.8$ & 	$\mathbf{69\%}$ & 	$\mathbf{8.8\%}$ 	\TBstrut\\
\Xhline{2\arrayrulewidth}

\end{tabular}
\end{center}
\end{footnotesize}
\label{Table_UTIn3D-H} 
\vspace{-3ex}
\end{table}

%
%
%
%
%
%

\subsection{SOGM Predictions in Real Scenarios}
\label{sec_real_preds}

In this section, we focus on the evaluation of the network SOGM predictions. Similarly to \cite{thomas2022learning}, we compare the predicted SOGMs to labeled SOGMs annotated by our automated pipeline, using the same metrics, and considering only the \textit{dynamic} class. In our first experiments shown in Table \ref{Table_lifelongsim}, we compare the performances of our network when trained on more and more data. The results are measured with mean Average Precision computed on the layer of SOGMs at 1, 2, and 3 seconds into the future. We also measure the total Average precision on the whole SOGM. The relatively low values in this table are explained by the complexity of the task. The future movements in the scene are never written in advance and can be multimodal (several possible trajectories are always possible). Therefore, the predictions incorporate this uncertainty and become blurry as time advances. It means lower precision scores, which reduces the values for our metric. The actual performances of our predictions can be judged with the qualitative visualizations we provide.

\begin{figure*}[b]
    \vspace{-1ex}
    \centering
    \includegraphics[width=0.999\textwidth, keepaspectratio=true]{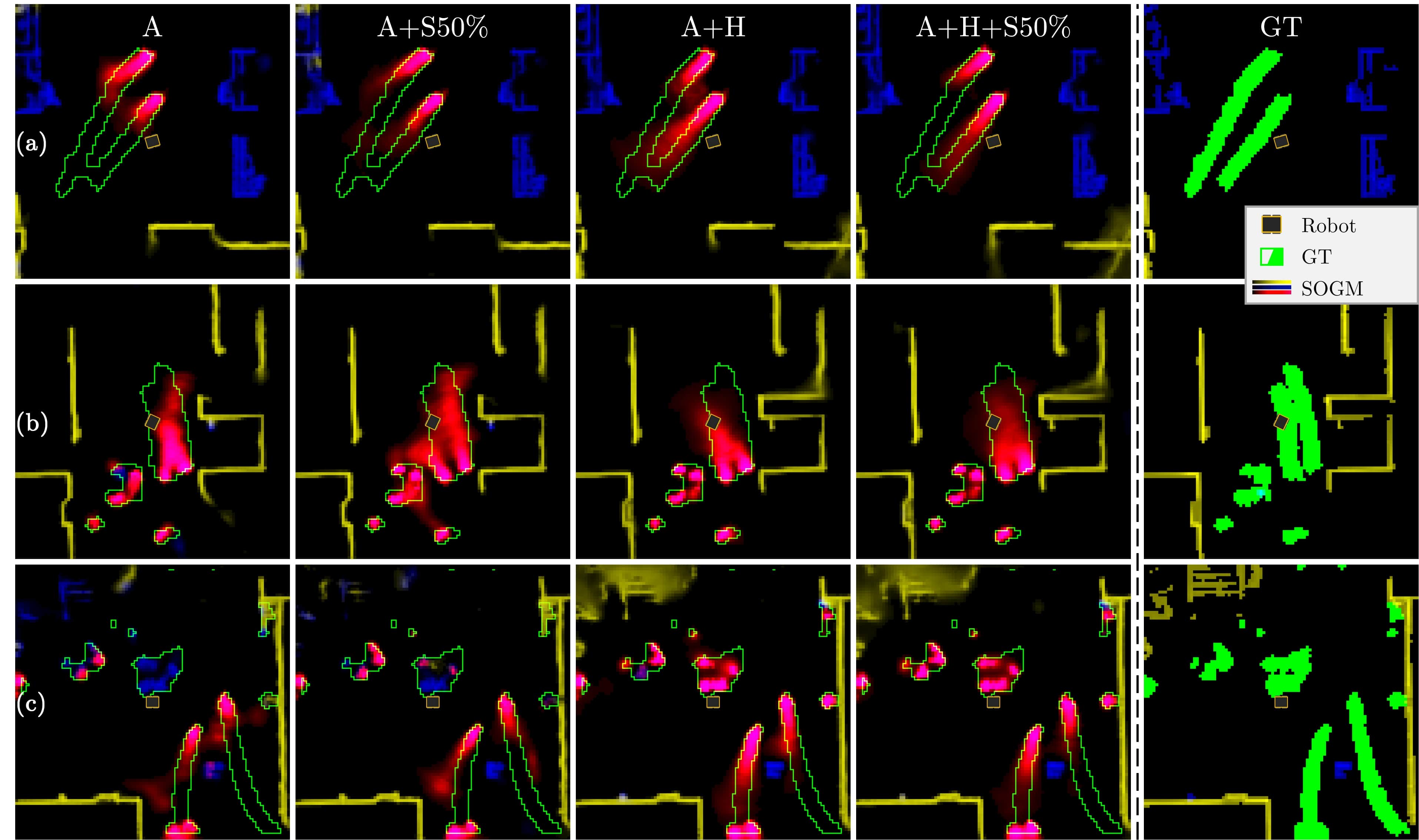}
    \caption{Qualitative comparison of SOGMs predicted with networks trained on an increasing amount of data. On all these examples from the UTIn3D-Hall dataset, we see that the more data we add to the training set, the better the predictions get. Our best network is trained on a training set combining UTIn3D-Atrium, UTIn3D-Hall, and simulated data.}
    \label{fig_sogmcomp}
\end{figure*}

First, in a lifelong learning manner, we use the UTIn3D-Atrium dataset, which has been split into 4 parts, and we train networks on increasing amounts of the dataset. We test on both UTIn3D validation sets, but we value the validation more in UTIn3D-Hall, as it contains a lot of "in-the-wild" dynamic obstacles. For this experiment, we see that increasing the amount and the diversity of data helps the network achieve better performances, which validates our assumption that a robot using our algorithm would be able to improve itself throughout its life. Because for other experiments, we used simulated data, we have a good opportunity to test the ability of our network to generalize to combinations of real-world and simulated data. We first notice that using only simulated data, the predictions are useless. Even if our simulated space is a copy of the UTIn3D-Atrium space, Sim-to-Real transfer is a very complex problem that we do not expect to solve here. However, when combining simulation data with real data in the training set, we see that the results are improved. The more we add simulated data to UTIn3D-Atrium, the better the performances are. It means, that without seeing any real data, the network is not able to generalize to this new unseen modality, but when given a few examples of the real data, the network can leverage the diversity of dynamic obstacles in the simulated data, at the same time as the specificity of the real data, to improve its performances. We believe that this is due to the fact that UTIn3D-Atrium does not contain a lot of dynamic obstacles (as shown in Table \ref{Table_UTIn3D-A}). The network only needs some frames of the dataset to adapt to the particularity of real data and get better results when it relies more on the movements seen in the simulated data, with many more actors.

\begin{table}[t]
\caption{Evaluation of the SOGM predictions with an increasing amount of data, and with combinations of data collected in the real world and simulated spaces. We provide the mean Average Precision (\%) at given future times (1s, 2s, and 3s) and on the whole SOGM (Total). Best results are highlighted in \textbf{bold} and results with 10\% of the best ones are \underline{underlined}.}
\setlength\tabcolsep{0.5pt}
\begin{footnotesize}
\begin{center}
\begin{tabular}{ L{1.79cm}  C{0.8cm} C{0.8cm} C{0.8cm} C{0.8cm} C{0.29cm} C{0.8cm} C{0.8cm} C{0.8cm} C{0.8cm} }

\multicolumn{1}{c}{} & \multicolumn{4}{c}{\textbf{UTIn3D-A-val}} & & \multicolumn{4}{c}{\textbf{UTIn3D-H-val}}  \\
\Xhline{2\arrayrulewidth}
Metrics & 1s & 2s & 3s & \textbf{Total}  & & 1s & 2s & 3s & \textbf{Total} \TBstrut\\
\Xhline{2\arrayrulewidth}

only-S	 & $3.0$	 & $0.4$	 & $0.1$	 & $4.2$	 &  & $8.7$	 & $1.2$	 & $0.4$	 & $8.5$	\TBstrut\\
\hline
A  (1)	 & $11.9$	 & $5.7$	 & $3.3$	 & $9.8$	 &  & $23.3$	 & $11.2$	 & $8.5$	 & $20.3$	\Tstrut\\
A  (12)	 & $13.9$	 & $5.9$	 & $3.9$	 & $10.6$	 &  & $29.9$	 & $11.5$	 & $6.9$	 & $21.9$	\\
A  (123)	 & $16.3$	 & $6.2$	 & $3.0$	 & $11.1$	 &  & $33.8$	 & $10.7$	 & $4.9$	 & $22.0$	\\
A  (1234)	 & $17.3$	 & $6.3$	 & $3.5$	 & $11.6$	 &  & $35.8$	 & $14.1$	 & $6.6$	 & $23.6$	\Bstrut\\
\hline
A+S20\%	 & $18.4$	 & $7.0$	 & $3.6$	 & $12.1$	 &  & $40.5$	 & $17.0$	 & $7.8$	 & $26.2$	\Tstrut\\
A+S50\%	 & $18.7$	 & $8.2$	 & $5.0$	 & $13.4$	 &  & $37.6$	 & $14.8$	 & $7.5$	 & $24.8$	\\
A+S80\%	 & $20.3$	 & $10.2$	 & $7.0$	 & $15.3$	 &  & $37.9$	 & $19.2$	 & $12.6$	 & $27.7$	\Bstrut\\
\hline
A+H	 & $\underline{24.2}$	 & $11.7$	 & $7.4$	 & $16.5$	 & & $\mathbf{56.9}$	 & $\underline{34.7}$	 & $\underline{24.5}$	 & $\underline{41.7}$	\TBstrut\\
\hline
A+H+S20\%	 & $\mathbf{26.7}$	 & $\mathbf{14.8}$	 & $\mathbf{9.5}$	 & $\mathbf{19.1}$	 &  & $\underline{56.4}$	 & $\underline{34.9}$	 & $\underline{24.7}$	 & $\underline{41.8}$	\Tstrut\\
A+H+S50\%	 & $\underline{24.7}$	 & $\underline{13.6}$	 & $\underline{8.6}$	 & $\underline{18.0}$	 &  & $\underline{56.5}$	 & $\mathbf{35.6}$	 & $\mathbf{25.6}$	 & $\mathbf{42.3}$	\\
A+H+S80\%	 & $21.3$	 & $10.6$	 & $7.0$	 & $15.6$	 &  & $\underline{54.1}$	 & $\underline{33.8}$	 & $\underline{24.8}$	 & $\underline{40.9}$	\Bstrut\\

\Xhline{2\arrayrulewidth}

\end{tabular}
\end{center}
\end{footnotesize}
\label{Table_lifelongsim}
\vspace{-3ex}
\end{table}

Then we also add UTIn3D-Hall to the training data. We notice a big step up in the performances on both validation sets. Following our analysis of the combined results with simulation, it makes sense, because UTIn3D-Hall contains a lot of dynamic obstacles (as shown in Table \ref{Table_UTIn3D-H}), with diverse behaviors. We notice that by combining both real training datasets, we achieve much better results than by combining one real dataset and simulated data. We could expect this, because, like the simulated data, UTIn3D-Hall contains a lot of dynamic obstacles, but it is not very different in nature from UTIn3D-Atrium. It is interesting to note that the gap between simulation and reality affects the performance more than the gap between two different spaces with different room configurations. This result confirms that our network does not overfit the training data and that its predicted SOGMs can generalize well to multiple different spaces.

Eventually, we combine everything, achieving the best results of all on both validation sets. This final result finally confirms that our network has the ability to generalize to combinations of diverse real-world and simulated spaces. The more data we provide, the better the results become, which is exactly the goal of our approach, as the robot should be able to collect this data on its own throughout its life. Interestingly, when we add more simulated data to the training set, we see the opposite as before: a reduction in the performance. In this case, when the real dataset is large enough, we think that adding some examples of simulated data helps by providing more diversity, but relying too much on it makes the network worse on real validation.

\begin{figure*}[t]
    \vspace{-1ex}
    \centering
    \includegraphics[width=0.999\textwidth, keepaspectratio=true]{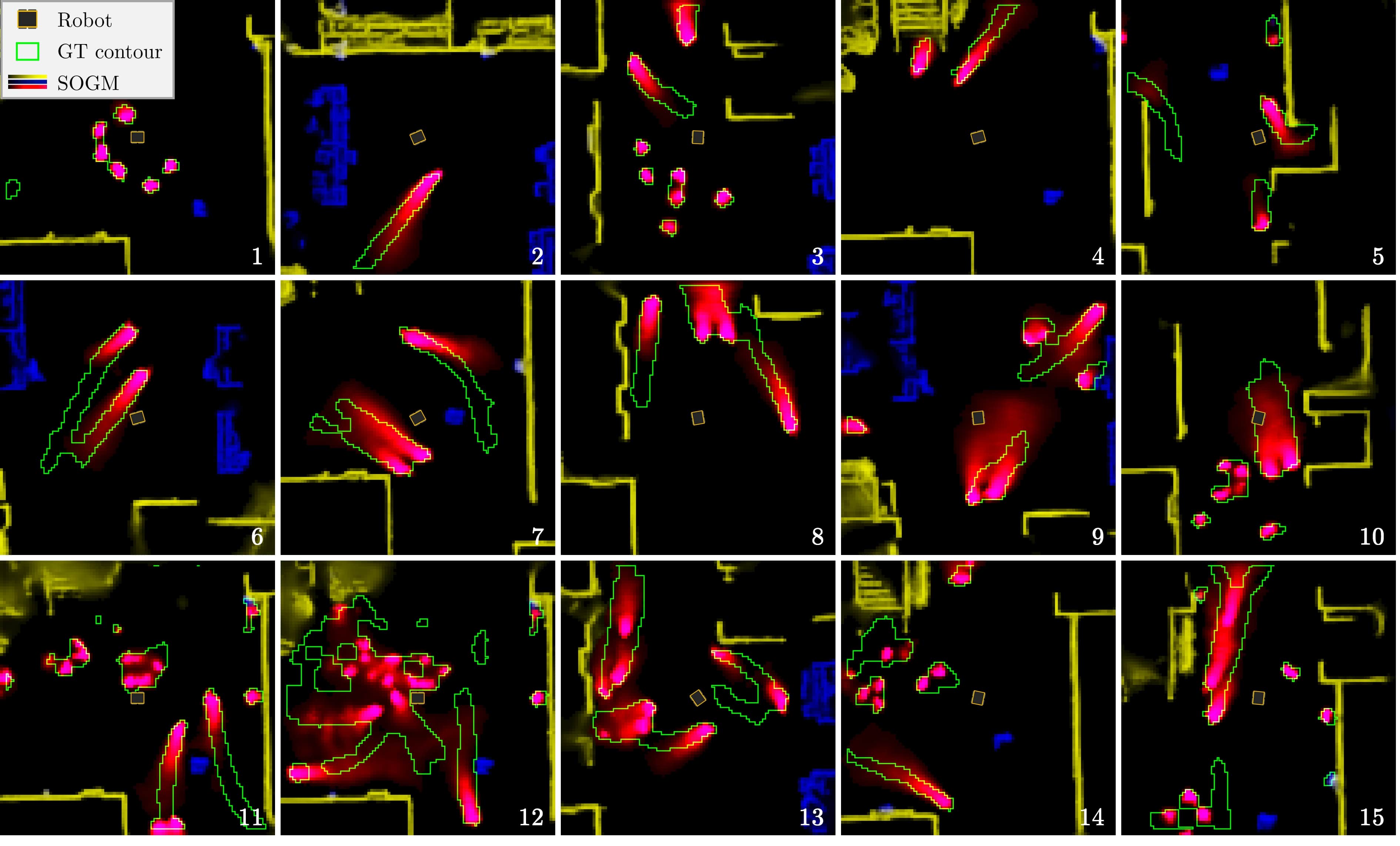}
    \caption{Examples of in-the-wild SOGM predictions with our best network (trained on A+H+S50\%), chosen in the UTIn3D-Hall validation set. Our network can handle various circumstances, from simple behaviors, like people standing around the robot, to extremely crowded and dynamic scenes.}
    \label{fig_sogmex}
\end{figure*}

We complete these quantitative results with qualitative examples of SOGM prediction for different training sets in Figure \ref{fig_sogmcomp}. Similarly to \cite{thomas2022learning}, we use a merged representation of the SOGM, where \textit{dynamic} predictions from all layers are colored in red, with the corresponding labeled SOGM superimposed as a green contour. The better performances of the network trained on A+H+S50\% can be visualized in all three examples. In example (a), we see two people walking in relatively free space. Using only A or A+S50\% data, the network is not able to predict the trajectories very well, as it has never seen the Hall. When using A+H, the trajectory gets better and takes the shape of a banana distribution, a phenomenon we previously saw in simulation \cite{thomas2022learning}. Finally, adding simulation with A+H+S50\%, helps refine the predictions, with a banana shape much closer to the green contour. In example (b), we notice that without simulation (A and A+H),  the predictions tend to merge for groups like these two persons. With simulation data (A+S50\% and A+H+S50\%), which contains a lot of examples with multiple actors, the predictions for groups get better. Eventually, in example (c) we notice groups of standing people that are classified as movable objects by the first networks and eventually classified as dynamic with A+H+S50\%. We do not consider this as an improvement, but more as an open question: should standing people be classified as movable or dynamic? Here this is decided by the network itself, depending on the data it has seen. In UTIn3D-Atrium data, people are standing for long periods, as they are discussing in the context of a conference, in UTIn3D-Hall, with a lot of passage, people are usually stopping only briefly, but then moving again, which explains the difference in the predictions.

In addition, more examples of predictions from A+H+S50\% are shown in \ref{fig_sogmex} to visualize the capabilities of our best network on real data. We show simple examples with only a few people (1-5), multiple people walking together as groups (6-10), and complex crowded scenes (11-15). Among these examples, we can find many interesting predictions. First, we notice the banana shape distribution (better seen as an animated SOGMs in the video) for one person but also groups (2, 4, 6-10, 14). We also see predictions of people fading when they get into elevators or stairs (4, 5, 13), or predictions expecting people to avoid the robot (5, 7, 9, 10).

%
%
%
%
%
%

\subsection{Real Robot Navigation}

The final step in our work is to test our navigation system in the real world, to validate the results we had in Section \ref{sec_simu} in simulation. In this section, we collect qualitative results and anecdotal examples, because we cannot reproduce a fair and accurate experimental process matching the one we have in simulation. We do not have groundtruth information that the simulator would provide, and more importantly, it is very hard to reproduce multiple experiments with the same conditions to compare different methods. 

\begin{table}[b]
\vspace{-3ex}
\caption{Evaluation of the safety and efficiency of our navigation systems with different predictions. We conduct two sessions with each system. Best results are highlighted in \textbf{bold} and results with 10\% of the best ones are \underline{underlined}.}
\setlength\tabcolsep{0.5pt}
\begin{footnotesize}
\begin{center}
\begin{tabular}{ L{1.95cm}  M{1.38cm} M{1.38cm} M{1.18cm} M{1.38cm} M{1.38cm}}

\Xhline{2\arrayrulewidth}
Nav System & Low-Risk Ratio (<$0.6$m) &    High-Risk Ratio (<$1.5$m) &	Time to Finish (s) &	Encounter min-dist (m, avg) & Encounter {\color{white}-}duration{\color{white}-} (s, avg) \TBstrut\\
\Xhline{2\arrayrulewidth}

\multirow{2}{*}{\textit{DeepSOGM-Nav}} 	 & $17.2\%$	 & $\mathbf{0.2\%}$	 & $\underline{91.7}$	 & $\mathbf{0.84}$	 & $\mathbf{1.99}$	\Tstrut\\
	 & $\mathbf{15.6\%}$	 & $1.9\%$	 & $\underline{90.5}$	 & $0.71$	 & $\underline{2.03}$	\Bstrut\\
\hline
\multirow{2}{*}{\textit{NoPreds-Nav}} 	 & $\underline{16.9\%}$	 & $2.8\%$	 & $105.1$	 & $0.58$	 & $2.24$	\Tstrut\\
	 & $\underline{16.3\%}$	 & $4.3\%$	 & $103.4$	 & $0.51$	 & $2.43$	\Bstrut\\
\hline
\textit{DeepSOGM-Nav}	 & $21.8\%$	 & $5.4\%$	 & $\underline{88.7}$	 & $0.51$	 & $2.42$	\Tstrut\\
\textit{(no time diff)}	 & $20.4\%$	 & $3.4\%$	 & $\mathbf{86.4}$	 & $0.53$	 & $2.21$	\Bstrut\\

\Xhline{2\arrayrulewidth}

\end{tabular}
\end{center}
\end{footnotesize}
\label{Table_realex}
\vspace{-3ex}
\end{table}

First, we conduct a controlled experiment where people are asked to cross the path of the robot perpendicularly and compare the reactions of the robot when using the standard \textit{NoPreds-Nav} or when using our \textit{DeepSOGM-Nav} with and without time diffusion. For this experiment, we repeated the session two times for each system and collected the data. After annotating it, we could collect the distance to the closest dynamic obstacle and compute similar metrics to the ones used in the simulated experiments in Section \ref{sec_simu}. We use different thresholds for the risk ratios, adapted to the real setup we defined: a Low-Risk Ratio measuring the proportion of the session during which the distance is smaller than $1.5m$), and a High-Risk Ratio measuring the proportion of the session during which the distance is smaller than $0.6m$). We also add two more metrics based on encounter statistics. For this, we segment out each encounter between the robot and a dynamic obstacle as the period when the distance is smaller than $1.5m$. For each encounter we measure the duration and the minimum distance, then we average these values per session.

In Table \ref{Table_realex}, we find that \textit{DeepSOGM-Nav} has the best performance in terms of safety and efficiency. It is very good at avoiding high-risk areas and keeping a higher minimum distance when crossing the path of people. We notice that without time diffusion for the SRM, \textit{DeepSOGM-Nav} is faster but a lot riskier. Overall, even though we cannot ensure the exact same conditions every time, get enough repetitions for a good statistical evaluation, and get groundtruth distance measurements; we still obtain similar results as in the simulation experiment.

\begin{figure}[t]
    \centering
    \includegraphics[width=0.999\columnwidth, keepaspectratio=true]{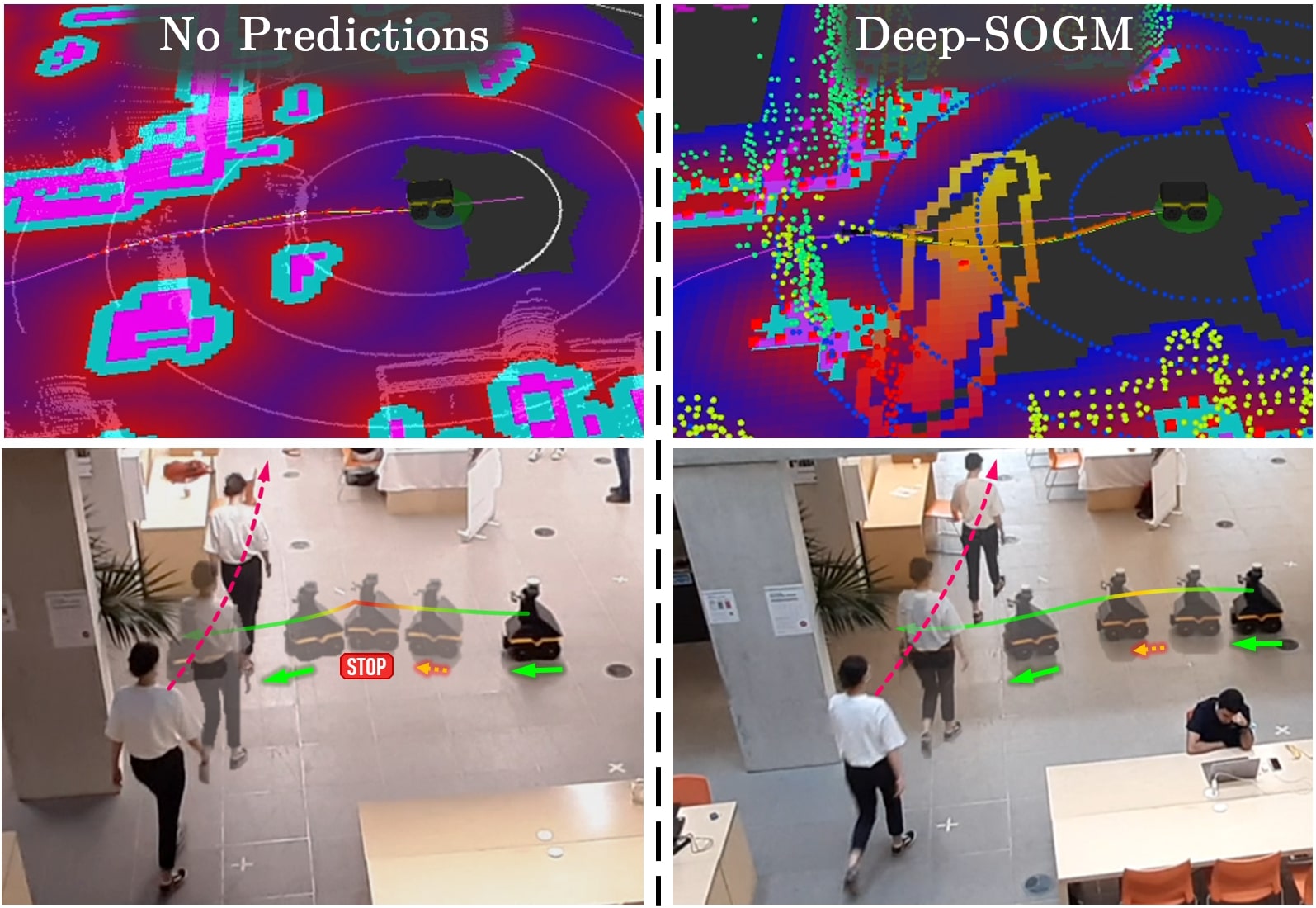}
    \caption{Anecdotal example showing the benefit of our Deep-SOGM predictions on a real robot. Without predictions, the robot sees an obstacle on its left and plans to go on the right. But the person is walking in that direction, forcing the robot to stop. With our predictions, the robot anticipates the person's movement and makes a plan to yield, without having to stop.}
    \label{fig_realex}
    \vspace{-3ex}
\end{figure}

We also show what happens qualitatively during these experiments in Figure \ref{fig_realex}. We see that without predictions, the planner sees an object coming from its left, and plans to avoid it by turning to the right. The closer this object gets, the further to the right the planned trajectory is ``pushed'', until the point where the robot has to stop and readjust its trajectory to pass on the left behind the person. By doing this the robot gets into a risky situation. On the contrary, with predictions, the planner anticipates that the person is going to be on its right in a few seconds, and, from the beginning, plans a trajectory that passes behind the person, which is much safer and more efficient. The reactions of the robot when using predictions are much more similar to what normal persons would do when crossing each others' paths.



%
%
%
%
%
%
%
%
%
%
%
%

\section{Conclusion}

In this paper, we presented a self-supervised approach that provides a robot with the ability to anticipate future movements in a dynamic scene. It can be seen as an imperfect but efficient crystal ball that does not rely on any human annotation. To gain this ability, a robot only needs to navigate in dynamic scenes, and our automated annotation pipeline will create a training set from the collected data. Our network architecture can then be trained to predict Spatiotemporal Occupancy Grid Maps, which contain information about the future of dynamic scenes. Finally, the robot can use this network within a simple navigation system, to get this information in real-time, transform it into Spatiotemporal Risk Maps, and use it in a local planner able to avoid high-risk areas in space and time.

Adapted from our previous work that was only tested on simulated data, our pipeline has been heavily improved, and thoroughly tested on real data. We compared our navigation pipeline with different kinds of predictions in simulation, validated the results on a real robot, and showed compelling SOGM prediction results in various circumstances. In addition, we provide a new 3D lidar dataset with indoor pedestrians, which contains lidar frames, with our annotations computed automatically.  This dataset can be used to reproduce our results or explore new potential methods for future predictions in dynamic scenes.

%
%
%
%
%
%
%
%
%
%
%
%




    



\bibliographystyle{IEEEtran}
\bibliography{IEEEabrv, TRO}

\begin{thebibliography}{10}
\providecommand{\url}[1]{#1}
\csname url@samestyle\endcsname
\providecommand{\newblock}{\relax}
\providecommand{\bibinfo}[2]{#2}
\providecommand{\BIBentrySTDinterwordspacing}{\spaceskip=0pt\relax}
\providecommand{\BIBentryALTinterwordstretchfactor}{4}
\providecommand{\BIBentryALTinterwordspacing}{\spaceskip=\fontdimen2\font plus
\BIBentryALTinterwordstretchfactor\fontdimen3\font minus
  \fontdimen4\font\relax}
\providecommand{\BIBforeignlanguage}[2]{{%
\expandafter\ifx\csname l@#1\endcsname\relax
\typeout{** WARNING: IEEEtran.bst: No hyphenation pattern has been}%
\typeout{** loaded for the language `#1'. Using the pattern for}%
\typeout{** the default language instead.}%
\else
\language=\csname l@#1\endcsname
\fi
#2}}
\providecommand{\BIBdecl}{\relax}
\BIBdecl

\bibitem{thomas2021self}
H.~Thomas, B.~Agro, M.~Gridseth, J.~Zhang, and T.~D. Barfoot, ``Self-supervised
  learning of lidar segmentation for autonomous indoor navigation,'' in
  \emph{2021 IEEE International Conference on Robotics and Automation
  (ICRA)}.\hskip 1em plus 0.5em minus 0.4em\relax IEEE, 2021.

\bibitem{thomas2022learning}
H.~Thomas, M.~G. d.~S. Aurin, J.~Zhang, and T.~D. Barfoot, ``Learning
  spatiotemporal occupancy grid maps for lifelong navigation in dynamic
  scenes,'' in \emph{2022 IEEE International Conference on Robotics and
  Automation (ICRA)}.\hskip 1em plus 0.5em minus 0.4em\relax IEEE, 2022.

\bibitem{ziebart2009planning}
B.~D. Ziebart, N.~Ratliff, G.~Gallagher, C.~Mertz, K.~Peterson, J.~A. Bagnell,
  M.~Hebert, A.~K. Dey, and S.~Srinivasa, ``Planning-based prediction for
  pedestrians,'' in \emph{2009 IEEE/RSJ International Conference on Intelligent
  Robots and Systems}.\hskip 1em plus 0.5em minus 0.4em\relax IEEE, 2009, pp.
  3931--3936.

\bibitem{kitani2012activity}
K.~M. Kitani, B.~D. Ziebart, J.~A. Bagnell, and M.~Hebert, ``Activity
  forecasting,'' in \emph{European Conference on Computer Vision}.\hskip 1em
  plus 0.5em minus 0.4em\relax Springer, 2012, pp. 201--214.

\bibitem{pomerleau2013comparing}
F.~Pomerleau, F.~Colas, R.~Siegwart, and S.~Magnenat, ``Comparing icp variants
  on real-world data sets,'' \emph{Autonomous Robots}, vol.~34, no.~3, pp.
  133--148, 2013.

\bibitem{zhang2014loam}
J.~Zhang and S.~Singh, ``Loam: Lidar odometry and mapping in real-time.'' in
  \emph{Robotics: Science and Systems}, vol.~2, no.~9, 2014.

\bibitem{mendes2016icp}
E.~Mendes, P.~Koch, and S.~Lacroix, ``Icp-based pose-graph slam,'' in
  \emph{2016 IEEE International Symposium on Safety, Security, and Rescue
  Robotics (SSRR)}.\hskip 1em plus 0.5em minus 0.4em\relax IEEE, 2016, pp.
  195--200.

\bibitem{deschaud2018imls}
J.-E. Deschaud, ``Imls-slam: scan-to-model matching based on 3d data,'' in
  \emph{2018 IEEE International Conference on Robotics and Automation
  (ICRA)}.\hskip 1em plus 0.5em minus 0.4em\relax IEEE, 2018, pp. 2480--2485.

\bibitem{moravec1985high}
H.~Moravec and A.~Elfes, ``High resolution maps from wide angle sonar,'' in
  \emph{Proceedings. 1985 IEEE international conference on robotics and
  automation}, vol.~2.\hskip 1em plus 0.5em minus 0.4em\relax IEEE, 1985, pp.
  116--121.

\bibitem{izadi2011kinectfusion}
S.~Izadi, D.~Kim, O.~Hilliges, D.~Molyneaux, R.~Newcombe, P.~Kohli, J.~Shotton,
  S.~Hodges, D.~Freeman, A.~Davison \emph{et~al.}, ``Kinectfusion: real-time 3d
  reconstruction and interaction using a moving depth camera,'' in
  \emph{Proceedings of the 24th annual ACM symposium on User interface software
  and technology}, 2011, pp. 559--568.

\bibitem{hornung2013octomap}
A.~Hornung, K.~M. Wurm, M.~Bennewitz, C.~Stachniss, and W.~Burgard, ``Octomap:
  An efficient probabilistic 3d mapping framework based on octrees,''
  \emph{Autonomous robots}, vol.~34, no.~3, pp. 189--206, 2013.

\bibitem{pomerleau2014long}
F.~Pomerleau, P.~Kr{\"u}si, F.~Colas, P.~Furgale, and R.~Siegwart, ``Long-term
  3d map maintenance in dynamic environments,'' in \emph{2014 IEEE
  International Conference on Robotics and Automation (ICRA)}.\hskip 1em plus
  0.5em minus 0.4em\relax IEEE, 2014, pp. 3712--3719.

\bibitem{biswas2014episodic}
J.~Biswas and M.~Veloso, ``Episodic non-markov localization: Reasoning about
  short-term and long-term features,'' in \emph{2014 IEEE International
  Conference on Robotics and Automation (ICRA)}.\hskip 1em plus 0.5em minus
  0.4em\relax IEEE, 2014, pp. 3969--3974.

\bibitem{dewan2017deep}
A.~Dewan, G.~L. Oliveira, and W.~Burgard, ``Deep semantic classification for 3d
  lidar data,'' in \emph{2017 IEEE/RSJ International Conference on Intelligent
  Robots and Systems (IROS)}.\hskip 1em plus 0.5em minus 0.4em\relax IEEE,
  2017, pp. 3544--3549.

\bibitem{sofman2006improving}
B.~Sofman, E.~Lin, J.~A. Bagnell, J.~Cole, N.~Vandapel, and A.~Stentz,
  ``Improving robot navigation through self-supervised online learning,''
  \emph{Journal of Field Robotics}, vol.~23, no. 11-12, pp. 1059--1075, 2006.

\bibitem{lookingbill2007reverse}
A.~Lookingbill, J.~Rogers, D.~Lieb, J.~Curry, and S.~Thrun, ``Reverse optical
  flow for self-supervised adaptive autonomous robot navigation,''
  \emph{International Journal of Computer Vision}, vol.~74, no.~3, pp.
  287--302, 2007.

\bibitem{hadsell2009learning}
R.~Hadsell, P.~Sermanet, J.~Ben, A.~Erkan, M.~Scoffier, K.~Kavukcuoglu,
  U.~Muller, and Y.~LeCun, ``Learning long-range vision for autonomous off-road
  driving,'' \emph{Journal of Field Robotics}, vol.~26, no.~2, pp. 120--144,
  2009.

\bibitem{brooks2012self}
C.~A. Brooks and K.~Iagnemma, ``Self-supervised terrain classification for
  planetary surface exploration rovers,'' \emph{Journal of Field Robotics},
  vol.~29, no.~3, pp. 445--468, 2012.

\bibitem{ridge2015self}
B.~Ridge, A.~Leonardis, A.~Ude, M.~Deni{\v{s}}a, and D.~Sko{\v{c}}aj,
  ``Self-supervised online learning of basic object push affordances,''
  \emph{International Journal of Advanced Robotic Systems}, vol.~12, no.~3,
  p.~24, 2015.

\bibitem{nava2019learning}
M.~Nava, J.~Guzzi, R.~O. Chavez-Garcia, L.~M. Gambardella, and A.~Giusti,
  ``Learning long-range perception using self-supervision from short-range
  sensors and odometry,'' \emph{IEEE Robotics and Automation Letters}, vol.~4,
  no.~2, pp. 1279--1286, 2019.

\bibitem{zhang2018semantic}
L.~Zhang, L.~Wei, P.~Shen, W.~Wei, G.~Zhu, and J.~Song, ``Semantic slam based
  on object detection and improved octomap,'' \emph{IEEE Access}, vol.~6, pp.
  75\,545--75\,559, 2018.

\bibitem{wang2019unified}
K.~Wang, Y.~Lin, L.~Wang, L.~Han, M.~Hua, X.~Wang, S.~Lian, and B.~Huang, ``A
  unified framework for mutual improvement of slam and semantic segmentation,''
  in \emph{2019 International Conference on Robotics and Automation
  (ICRA)}.\hskip 1em plus 0.5em minus 0.4em\relax IEEE, 2019, pp. 5224--5230.

\bibitem{chen2019suma++}
X.~Chen, A.~Milioto, E.~Palazzolo, P.~Gigu{\`e}re, J.~Behley, and C.~Stachniss,
  ``Suma++: Efficient lidar-based semantic slam,'' in \emph{2019 IEEE/RSJ
  International Conference on Intelligent Robots and Systems (IROS)}.\hskip 1em
  plus 0.5em minus 0.4em\relax IEEE, 2019, pp. 4530--4537.

\bibitem{sun2018recurrent}
L.~Sun, Z.~Yan, A.~Zaganidis, C.~Zhao, and T.~Duckett, ``Recurrent-octomap:
  Learning state-based map refinement for long-term semantic mapping with
  3-d-lidar data,'' \emph{IEEE Robotics and Automation Letters}, vol.~3, no.~4,
  pp. 3749--3756, 2018.

\bibitem{alahi2016social}
A.~Alahi, K.~Goel, V.~Ramanathan, A.~Robicquet, L.~Fei-Fei, and S.~Savarese,
  ``Social lstm: Human trajectory prediction in crowded spaces,'' in
  \emph{Proceedings of the IEEE conference on computer vision and pattern
  recognition}, 2016, pp. 961--971.

\bibitem{gupta2018social}
A.~Gupta, J.~Johnson, L.~Fei-Fei, S.~Savarese, and A.~Alahi, ``Social gan:
  Socially acceptable trajectories with generative adversarial networks,'' in
  \emph{Proceedings of the IEEE Conference on Computer Vision and Pattern
  Recognition}, 2018, pp. 2255--2264.

\bibitem{katyal2020intent}
K.~D. Katyal, G.~D. Hager, and C.-M. Huang, ``Intent-aware pedestrian
  prediction for adaptive crowd navigation,'' in \emph{2020 IEEE International
  Conference on Robotics and Automation (ICRA)}.\hskip 1em plus 0.5em minus
  0.4em\relax IEEE, 2020, pp. 3277--3283.

\bibitem{peddi2020data}
R.~Peddi, C.~Di~Franco, S.~Gao, and N.~Bezzo, ``A data-driven framework for
  proactive intention-aware motion planning of a robot in a human
  environment,'' in \emph{2020 IEEE/RSJ International Conference on Intelligent
  Robots and Systems (IROS)}.\hskip 1em plus 0.5em minus 0.4em\relax IEEE,
  2020, pp. 5738--5744.

\bibitem{sathyamoorthy2020frozone}
A.~J. Sathyamoorthy, U.~Patel, T.~Guan, and D.~Manocha, ``Frozone:
  Freezing-free, pedestrian-friendly navigation in human crowds,'' \emph{IEEE
  Robotics and Automation Letters}, vol.~5, no.~3, pp. 4352--4359, 2020.

\bibitem{luo2018fast}
W.~Luo, B.~Yang, and R.~Urtasun, ``Fast and furious: Real time end-to-end 3d
  detection, tracking and motion forecasting with a single convolutional net,''
  in \emph{Proceedings of the IEEE conference on Computer Vision and Pattern
  Recognition}, 2018, pp. 3569--3577.

\bibitem{casas2018intentnet}
S.~Casas, W.~Luo, and R.~Urtasun, ``Intentnet: Learning to predict intention
  from raw sensor data,'' in \emph{Conference on Robot Learning}.\hskip 1em
  plus 0.5em minus 0.4em\relax PMLR, 2018, pp. 947--956.

\bibitem{jain2020discrete}
A.~Jain, S.~Casas, R.~Liao, Y.~Xiong, S.~Feng, S.~Segal, and R.~Urtasun,
  ``Discrete residual flow for probabilistic pedestrian behavior prediction,''
  in \emph{Conference on Robot Learning}.\hskip 1em plus 0.5em minus
  0.4em\relax PMLR, 2020, pp. 407--419.

\bibitem{chen2017socially}
Y.~F. Chen, M.~Everett, M.~Liu, and J.~P. How, ``Socially aware motion planning
  with deep reinforcement learning,'' in \emph{2017 IEEE/RSJ International
  Conference on Intelligent Robots and Systems (IROS)}.\hskip 1em plus 0.5em
  minus 0.4em\relax IEEE, 2017, pp. 1343--1350.

\bibitem{long2018towards}
P.~Long, T.~Fan, X.~Liao, W.~Liu, H.~Zhang, and J.~Pan, ``Towards optimally
  decentralized multi-robot collision avoidance via deep reinforcement
  learning,'' in \emph{2018 IEEE International Conference on Robotics and
  Automation (ICRA)}.\hskip 1em plus 0.5em minus 0.4em\relax IEEE, 2018, pp.
  6252--6259.

\bibitem{liang2020crowdsteer}
J.~Liang, U.~Patel, A.~Sathyamoorthy, and D.~Manocha, ``Crowdsteer: Realtime
  smooth and collision-free robot navigation in densely crowded scenarios
  trained using high-fidelity simulation,'' in \emph{Proceedings of the
  Twenty-Ninth International Joint Conference on Artificial Intelligence,
  IJCAI}, vol. 2020, 2020, pp. 4221--4228.

\bibitem{sathyamoorthy2020densecavoid}
A.~J. Sathyamoorthy, J.~Liang, U.~Patel, T.~Guan, R.~Chandra, and D.~Manocha,
  ``Densecavoid: Real-time navigation in dense crowds using anticipatory
  behaviors,'' in \emph{2020 IEEE International Conference on Robotics and
  Automation (ICRA)}.\hskip 1em plus 0.5em minus 0.4em\relax IEEE, 2020, pp.
  11\,345--11\,352.

\bibitem{everett2021collision}
M.~Everett, Y.~F. Chen, and J.~P. How, ``Collision avoidance in pedestrian-rich
  environments with deep reinforcement learning,'' \emph{IEEE Access}, vol.~9,
  pp. 10\,357--10\,377, 2021.

\bibitem{strudel2020learning}
R.~Strudel, R.~Garcia, J.~Carpentier, J.-P. Laumond, I.~Laptev, and C.~Schmid,
  ``Learning obstacle representations for neural motion planning,'' 2020.

\bibitem{liu2020robot}
L.~Liu, D.~Dugas, G.~Cesari, R.~Siegwart, and R.~Dub{\'e}, ``Robot navigation
  in crowded environments using deep reinforcement learning,'' in \emph{2020
  IEEE/RSJ International Conference on Intelligent Robots and Systems
  (IROS)}.\hskip 1em plus 0.5em minus 0.4em\relax {IEEE}, 2020, pp. 5671--5677.

\bibitem{patel2021dwa}
U.~Patel, N.~K.~S. Kumar, A.~J. Sathyamoorthy, and D.~Manocha, ``Dwa-rl:
  Dynamically feasible deep reinforcement learning policy for robot navigation
  among mobile obstacles,'' in \emph{2021 IEEE International Conference on
  Robotics and Automation (ICRA)}.\hskip 1em plus 0.5em minus 0.4em\relax IEEE,
  2020.

\bibitem{pierson2019dynamic}
A.~Pierson, C.-I. Vasile, A.~Gandhi, W.~Schwarting, S.~Karaman, and D.~Rus,
  ``Dynamic risk density for autonomous navigation in cluttered environments
  without object detection,'' in \emph{2019 International Conference on
  Robotics and Automation (ICRA)}.\hskip 1em plus 0.5em minus 0.4em\relax IEEE,
  2019, pp. 5807--5814.

\bibitem{huang2020safe}
Z.~Huang, W.~Schwarting, A.~Pierson, H.~Guo, M.~Ang, and D.~Rus, ``Safe path
  planning with multi-model risk level sets,'' in \emph{2020 IEEE/RSJ
  International Conference on Intelligent Robots and Systems (IROS)}.\hskip 1em
  plus 0.5em minus 0.4em\relax IEEE, 2020, pp. 6268--6275.

\bibitem{fisac2018probabilistically}
J.~F. Fisac, A.~Bajcsy, S.~L. Herbert, D.~Fridovich-Keil, S.~Wang, C.~J.
  Tomlin, and A.~D. Dragan, ``Probabilistically safe robot planning with
  confidence-based human predictions,'' \emph{arXiv preprint arXiv:1806.00109},
  2018.

\bibitem{bajcsy2019scalable}
A.~Bajcsy, S.~L. Herbert, D.~Fridovich-Keil, J.~F. Fisac, S.~Deglurkar, A.~D.
  Dragan, and C.~J. Tomlin, ``A scalable framework for real-time multi-robot,
  multi-human collision avoidance,'' in \emph{2019 international conference on
  robotics and automation (ICRA)}.\hskip 1em plus 0.5em minus 0.4em\relax IEEE,
  2019, pp. 936--943.

\bibitem{bansal2020hamilton}
S.~Bansal, A.~Bajcsy, E.~Ratner, A.~D. Dragan, and C.~J. Tomlin, ``A
  hamilton-jacobi reachability-based framework for predicting and analyzing
  human motion for safe planning,'' in \emph{2020 IEEE International Conference
  on Robotics and Automation (ICRA)}.\hskip 1em plus 0.5em minus 0.4em\relax
  IEEE, 2020, pp. 7149--7155.

\bibitem{lotter2016deep}
W.~Lotter, G.~Kreiman, and D.~Cox, ``Deep predictive coding networks for video
  prediction and unsupervised learning,'' \emph{arXiv preprint
  arXiv:1605.08104}, 2016.

\bibitem{wang2018predrnn++}
Y.~Wang, Z.~Gao, M.~Long, J.~Wang, and S.~Y. Philip, ``Predrnn++: Towards a
  resolution of the deep-in-time dilemma in spatiotemporal predictive
  learning,'' in \emph{International Conference on Machine Learning}.\hskip 1em
  plus 0.5em minus 0.4em\relax PMLR, 2018, pp. 5123--5132.

\bibitem{wang2019memory}
Y.~Wang, J.~Zhang, H.~Zhu, M.~Long, J.~Wang, and P.~S. Yu, ``Memory in memory:
  A predictive neural network for learning higher-order non-stationarity from
  spatiotemporal dynamics,'' in \emph{Proceedings of the IEEE/CVF Conference on
  Computer Vision and Pattern Recognition}, 2019, pp. 9154--9162.

\bibitem{mohajerin2019multi}
N.~Mohajerin and M.~Rohani, ``Multi-step prediction of occupancy grid maps with
  recurrent neural networks,'' in \emph{Proceedings of the IEEE/CVF Conference
  on Computer Vision and Pattern Recognition}, 2019, pp. 10\,600--10\,608.

\bibitem{schreiber2020motion}
M.~Schreiber, V.~Belagiannis, C.~Gl{\"a}ser, and K.~Dietmayer, ``Motion
  estimation in occupancy grid maps in stationary settings using recurrent
  neural networks,'' in \emph{2020 IEEE International Conference on Robotics
  and Automation (ICRA)}.\hskip 1em plus 0.5em minus 0.4em\relax IEEE, 2020,
  pp. 8587--8593.

\bibitem{toyungyernsub2020double}
M.~Toyungyernsub, M.~Itkina, R.~Senanayake, and M.~J. Kochenderfer,
  ``Double-prong convlstm for spatiotemporal occupancy prediction in dynamic
  environments,'' in \emph{2021 International Conference on Robotics and
  Automation (ICRA)}.\hskip 1em plus 0.5em minus 0.4em\relax IEEE, 2021.

\bibitem{shoemake1985animating}
K.~Shoemake, ``Animating rotation with quaternion curves,'' in
  \emph{Proceedings of the 12th annual conference on Computer graphics and
  interactive techniques}, 1985, pp. 245--254.

\bibitem{matheron2002birth}
G.~Matheron and J.~Serra, ``The birth of mathematical morphology,'' in
  \emph{Proc. 6th Intl. Symp. Mathematical Morphology}.\hskip 1em plus 0.5em
  minus 0.4em\relax Sydney, Australia, 2002, pp. 1--16.

\bibitem{calderon2014point}
S.~Calderon and T.~Boubekeur, ``Point morphology,'' \emph{ACM Transactions on
  Graphics (TOG)}, vol.~33, no.~4, pp. 1--13, 2014.

\bibitem{balado2020mathematical}
J.~Balado, P.~Van~Oosterom, L.~D{\'\i}az-Vilari{\~n}o, and M.~Meijers,
  ``Mathematical morphology directly applied to point cloud data,'' \emph{ISPRS
  Journal of Photogrammetry and Remote Sensing}, vol. 168, pp. 208--220, 2020.

\bibitem{zhou2018open3d}
Q.-Y. Zhou, J.~Park, and V.~Koltun, ``Open3d: A modern library for 3d data
  processing,'' \emph{arXiv preprint arXiv:1801.09847}, 2018.

\bibitem{thomas2019kpconv}
H.~Thomas, C.~R. Qi, J.-E. Deschaud, B.~Marcotegui, F.~Goulette, and L.~J.
  Guibas, ``Kpconv: Flexible and deformable convolution for point clouds,'' in
  \emph{Proceedings of the IEEE/CVF International Conference on Computer
  Vision}, 2019, pp. 6411--6420.

\bibitem{rosmann2015planning}
C.~R{\"o}smann, F.~Hoffmann, and T.~Bertram, ``Planning of multiple robot
  trajectories in distinctive topologies,'' in \emph{2015 European Conference
  on Mobile Robots (ECMR)}.\hskip 1em plus 0.5em minus 0.4em\relax IEEE, 2015,
  pp. 1--6.

\bibitem{rosmann2017integrated}
------, ``Integrated online trajectory planning and optimization in distinctive
  topologies,'' \emph{Robotics and Autonomous Systems}, vol.~88, pp. 142--153,
  2017.

\end{thebibliography}

\begin{IEEEbiography}[{\includegraphics[width=1in,height=1.25in,clip,keepaspectratio]{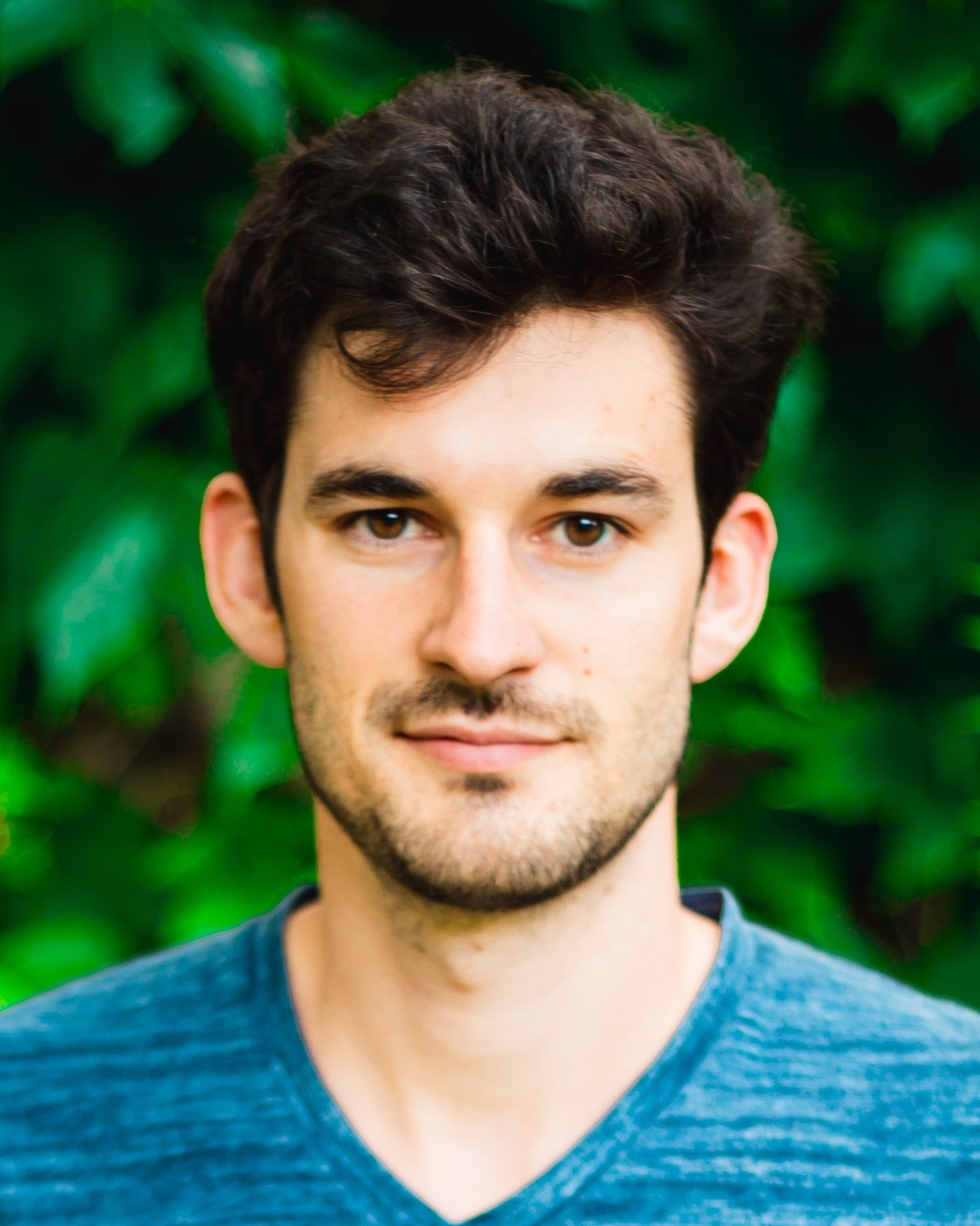}}]{Hugues Thomas} (Member, IEEE) received the Ph.D. degree from Mines Paristech, Université Paris Sciences et Lettres (PSL), Paris, France, in 2019. He is currently a Postdoctoral Researcher with the Autonomous Space Robotics Lab (ASRL), University of Toronto, which develops methods to allow mobile robots to operate in large-scale, unstructured, 3-D environments, using rich onboard sensing (e.g., cameras and laser rangefinders) and computation.

His research interests focus on deep learning, 3D point clouds and robotics. He has developed a novel convolutional operator for 3D point clouds, KPConv, designed deep network architectures for 3D object classification, object part segmentation and semantic scene parsing, explored the problems of rotation invariance and equivariance in 3D convolutions, and studied the application of self-supervised learning to robotics applications.

\end{IEEEbiography}

\begin{IEEEbiography}[{\includegraphics[width=1in,height=1.25in,clip,keepaspectratio]{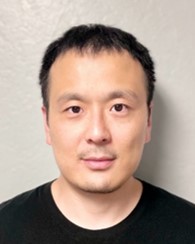}}]{Jian Zhang} (Member, IEEE) received the B.S. degree in mechatronics from Zhejiang University, Hangzhou, China, in 2010, and the Ph.D. degree in mechanical engineering with robotics specialty from Purdue University, West Lafayette, IN, USA, in 2015. He is currently an R\&D manager at Apple Inc., USA. His research interests are robotics, autonomous systems, deep learning, and embodied artificial intelligence.

\end{IEEEbiography}

\begin{IEEEbiography}[{\includegraphics[width=1in,height=1.25in,clip,keepaspectratio]{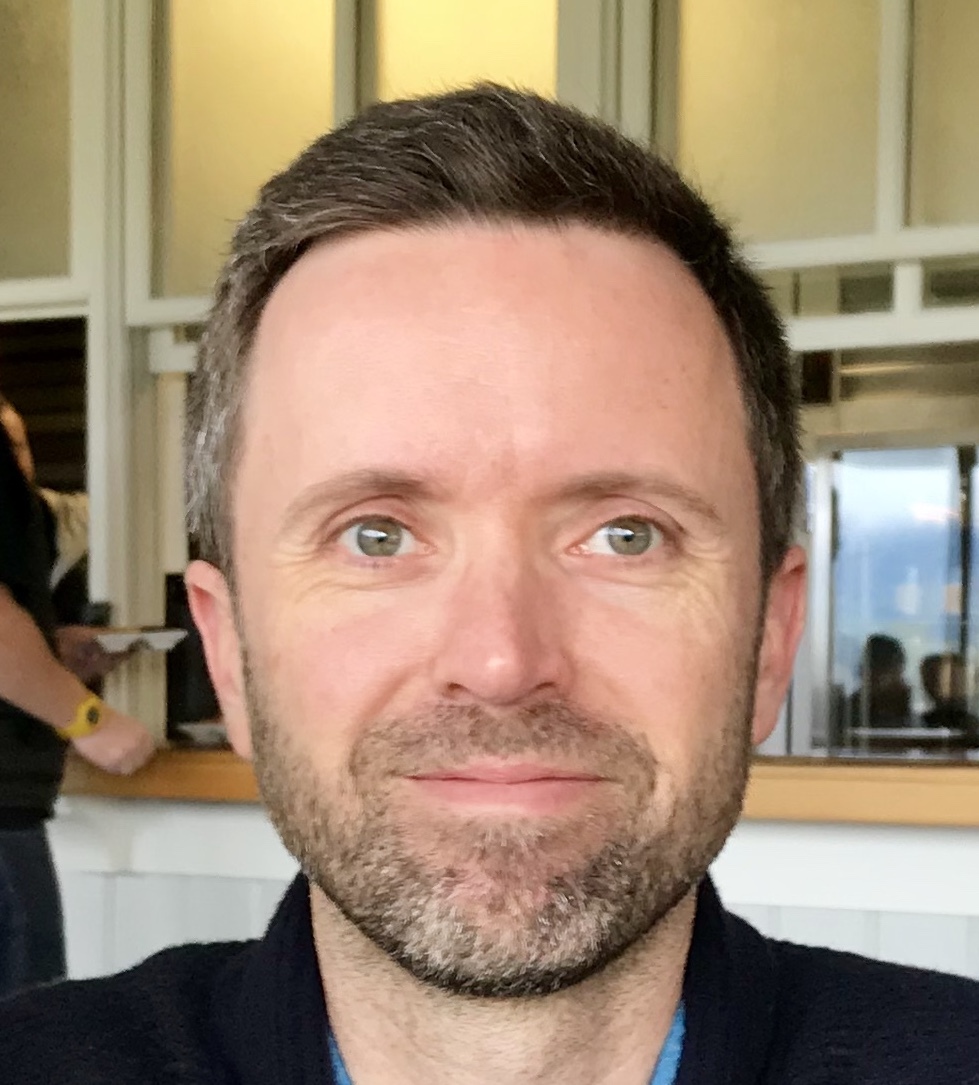}}]{Timothy D. Barfoot} (Fellow, IEEE) received the Ph.D. degree from University of Toronto, Toronto, ON, Canada, in 2002. He currently leads the Autonomous Space Robotics Lab (ASRL), University of Toronto, and works on the area of autonomy for mobile robots targeting a variety of applications. 

He held a Canada Research Chair (Tier 2) for the full 10 years and was an Early Researcher Awardee in the Province of Ontario. Timothy was also a Visiting Professor at the University of Oxford in 2013 and recently completed a leave as Director of Autonomous Systems at Apple in California in 2017-9. He is currently the Chair of the UofT Engineering Science Robotics Major, Associate Director of the UofT Robotics Institute, Faculty Affiliate of the Vector Institute, and co-Faculty Advisor of UofT's self-driving car team that won the SAE Autodrive competition five years in a row.  He sits on the Editorial Boards of the International Journal of Robotics Research (IJRR) and Field Robotics (FR), the Foundation Board of Robotics: Science and Systems (RSS), and served as the General Chair of Field and Service Robotics (FSR) 2015, which was held in Toronto. He is the author of a book, ``State Estimation for Robotics''.
\end{IEEEbiography}

\vfill

\end{document}